\PassOptionsToPackage{table,dvipsnames}{xcolor}
\documentclass{article}

\usepackage[preprint]{corl_2026} % Uncomment for pre-prints (e.g., arxiv); This is like ``final'', but will remove the CORL footnote.

\usepackage{hyperref}

\usepackage[utf8]{inputenc} % allow utf-8 input
\usepackage[T1]{fontenc}    % use 8-bit T1 fonts
\usepackage{url}            % simple URL typesetting
\usepackage{booktabs}       % professional-quality tables
\usepackage{amsfonts}       % blackboard math symbols
\usepackage{nicefrac}       % compact symbols for 1/2, etc.
\usepackage{microtype}      % microtypography
\usepackage{xcolor}         % colors

\usepackage{amssymb}
\usepackage{amsmath}
\usepackage{subcaption}
\usepackage{graphicx}
\usepackage{enumitem}
\usepackage{multirow}
\usepackage{amsthm}
\usepackage[linesnumbered,ruled,vlined]{algorithm2e}
\usepackage{float}
\usepackage{wrapfig}
\usepackage[commandnameprefix=always,authormarkup=none]{changes}

\definechangesauthor[name={Rebuttal integration}, color=blue]{rebuttal}

\usepackage{tikz}

\setlist{
    topsep=-3pt,        % space above/below list
    itemsep=2pt,       % space between items
    parsep=0pt,        % space between paragraphs within items
    partopsep=0pt,     % extra space when list starts a new paragraph
    leftmargin=*,      % reduce indentation automatically
}

\definecolor{lightgray}{gray}{0.95}

\title{What Matters in Orchestrating Robot Policies: \\A Systematic Study of Hierarchical VLA Agents}
\author{
  Jiaheng Hu\thanks{Work done while interning at Google DeepMind}\hspace{0.4em}, Mohit Shridhar, Caden Lu, Dhruv Shah, \\
  \textbf{Hao-Tien Lewis Chiang, Jie Tan, Annie Xie}\\\\
  Google DeepMind\\
}

\begin{document}
\maketitle

%===============================================================================


%===============================================================================
\begin{abstract}
Hierarchical vision-language-action (Hi-VLA) systems have emerged as a promising paradigm for complex robot manipulation, by using high-level VLM planners to decompose tasks into language subgoals executed by low-level VLA controllers. Despite recent empirical progress, there is a lack of unified design principles for these systems: existing Hi-VLA systems differ in how they choose and connect planners, controllers, mechanisms to switch between the two, and how observations and memory are represented in the planner. In this paper, we present a systematic study of Hi-VLA design for robot manipulation. We unify representative Hi-VLA agents under an options-style control framework and benchmark core design choices across short-horizon, long-horizon, and reasoning-intensive tasks. Our analysis distills practical principles for building Hi-VLA systems, showing how model choices and interface mechanisms jointly shape performance. Applying these principles yields a substantially stronger system than either flat VLA control or a naively designed hierarchy, across experiments both in simulation and on a real ALOHA robot.
Overall, our results provide a foundation for building more capable, robust, and principled hierarchical VLA agents. More information and video at \url{jiahenghu.github.io/hi-vla}.
\end{abstract}

% \keywords{Hierarchical VLA, Robot Manipulation}     
\section{Introduction}
Recent advances in vision-language-action (VLA) models~\cite{team2025gemini, black2024pi_0, kim2024openvla,zitkovich2023rt,wen2025tinyvla, abdolmaleki2025gemini,o2024open, nvidia2025gr00tn1openfoundation} have demonstrated impressive generalization capabilities in solving robotics tasks by directly mapping natural-language commands to robot actions. These models offer a high degree of steerability, enabling robots to execute open-ended and potentially nuanced language prompts, such as \emph{``put the red mug on top of the blue plate on the left.''}
% By fine-tuning large vision-language models (VLMs) \cite{beyer2024paligemma, team2023gemini,bai2023qwen,zhu2023minigpt} on robot data~\cite{o2024open,khazatsky2024droid}, systems such as Gemini Robotics~\cite{team2025gemini}, OpenVLA~\cite{kim2024openvla}, and Pi~\cite{black2024pi_0} can directly map natural-language commands to robot actions, representing a promising step toward the creation of generalist robot agents. 
However, monolithic VLAs still remain limited in their ability to perform long-horizon, compositional, or abstract reasoning tasks~\cite{black2024pi_0, team2025gemini}, due to two main reasons. First, since these models are primarily trained with short, easy-to-collect trajectory segments, they struggle to generalize to long-horizon tasks and commands. Second, fine-tuning VLMs on action data often catastrophically compromises the VLM’s original reasoning and compositional capabilities, preventing them from repurposing learned robot skills towards novel or complex tasks.

% Such a deficiency stands in stark contrast to the core capability of vision-language models (VLMs) themselves. When prompted, VLMs have demonstrated impressive aptitude for abstract reasoning and complex common-sense planning, especially through the use of explicit chains of thought ($\text{CoT}$) \cite{wei2022chain, driess2023palm, ahn2022can, huang2022inner}. This capability gap motivates the creation of \textbf{hierarchical VLA systems} (Fig.~\ref{fig:pull})~\cite{belkhale2024rt,intelligence2025pi_,abdolmaleki2025gemini,shi2025hi,li2025hamster,figure2024helix}.
% Partially inspired by Daniel Kahneman’s famous “Thinking, fast and slow” framework\cite{kahneman2011thinking}, these hierarchical systems employ a high-level VLM planner to decompose tasks into sub-goals that are executed by low-level VLAs, thereby naturally combine the reasoning ability of VLMs with the fine-grained control of robot policies.  

\begin{figure*}
    \vspace{-0.2cm}
    \centering
    \includegraphics[width=0.99\linewidth]{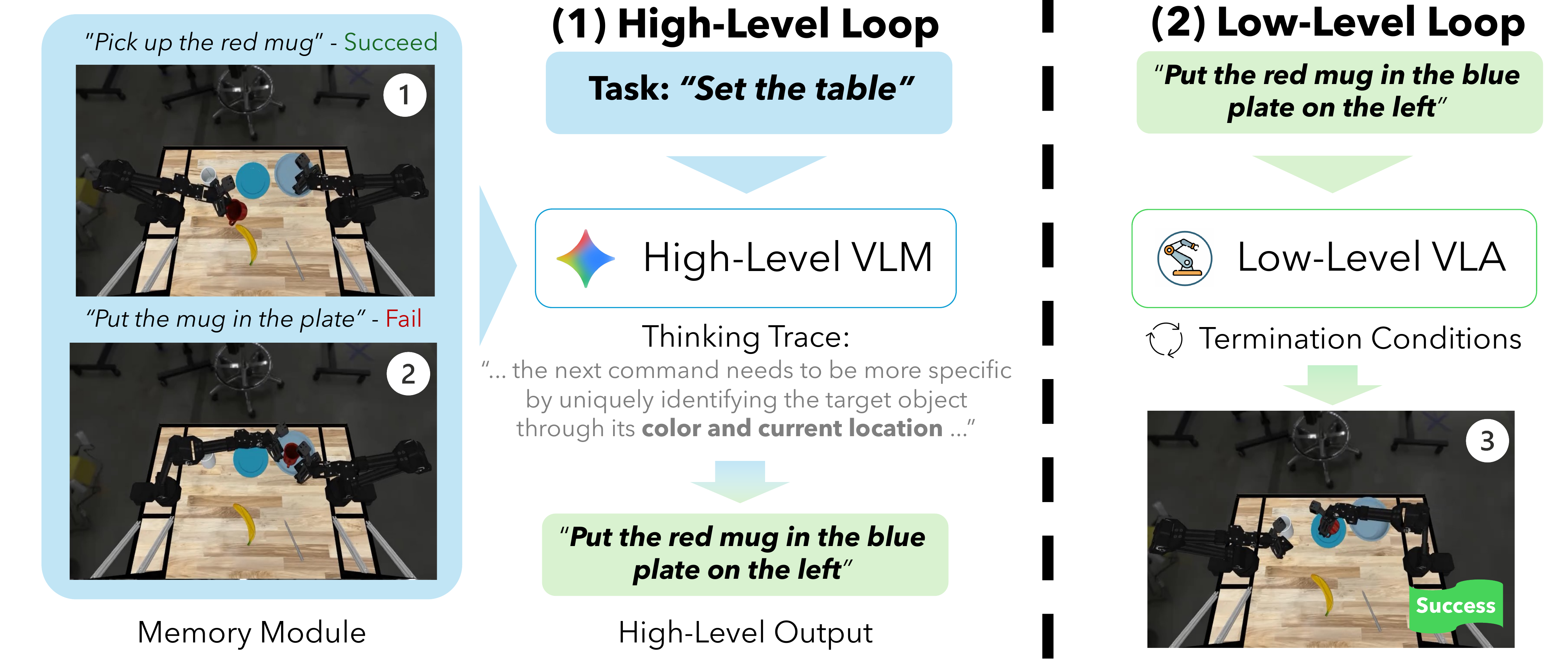}
    \caption{\small Hierarchical VLA systems have the potential to compensate for the deficiencies of the low-level VLA by generating suitable commands, thereby achieving compositional generalization, especially for long-horizon and reasoning tasks. In this paper, we study the key design choices of Hi-VLA systems, towards a better understanding of how and why they impact the overall performance.}
    \label{fig:pull}
    \vspace{-0.3cm}
\end{figure*}

Hierarchical VLA systems (Fig.~\ref{fig:pull})~\cite{belkhale2024rt,intelligence2025pi_,abdolmaleki2025gemini,shi2025hi,li2025hamster,figure2024helix} naturally address this challenge by introducing a high-level VLM planner that reasons over the task and scene, proposes language subgoals, and delegates execution to a low-level VLA policy. This biologically inspired division of labor~\cite{kahneman2011thinking} allows the system to combine the semantic and compositional strengths of VLMs with the physical grounding of VLAs. Recent Hi-VLA systems can perform multi-step household tasks~\cite{team2025gemini,intelligence2025pi_}, adapt across embodiments~\cite{tan2025roboos}, and perform semantic reasoning \cite{belkhale2024rt}, suggesting hierarchy can be a powerful paradigm for more capable embodied agents.

Despite these successes, the design principles behind Hi-VLA systems remain under-explored. 
A Hi-VLA agent is a complex system shaped by many design choices: the choice of VLM planner and the low-level VLA, the observation representation given to the VLM, the criteria for switching control back from the VLA to the VLM, and the memory mechanism. Existing systems instantiate these choices in different ways, making it difficult to determine which components are important and which design choices are most relevant for different task regimes.

In this work, we take a step toward a principled understanding of hierarchical VLA design. We first unify hierarchical VLA agents under a shared control loop inspired by the \textit{options framework}~\cite{sutton1999between}, which allows us to isolate and evaluate major design choices in a controlled manner. 
% In this formulation, the high-level VLM acts as an option-selection policy that generates temporally extended language commands, while the low-level VLA acts as an intra-option policy that executes those commands through robot actions. The observation representation, memory module, and termination condition mediate the interface between these two policies. 
Based on this framework, we conduct a systematic empirical study across diverse manipulation tasks both in simulation and in the real world, spanning three task categories: short-horizon tasks that resemble the length of typical VLA training trajectories, long-horizon tasks that require composing multiple short-horizon skills, and reasoning tasks that require interpreting indirect or semantic instructions. This structured evaluation allows us to ask not only whether a design choice improves average success, but also where it matters most: atomic execution, skill composition, or semantic reasoning.

Our results reveal that while a naive Hi-VLA can improve over flat VLA, carefully chosen hierarchical designs yield substantially larger gains, especially on long-horizon and reasoning-intensive tasks. Specifically, strong Hi-VLA performance depends jointly on the model backbones and the interfaces between them: reasoning-enabled VLMs improve high-level decision-making, steerable VLA controllers are essential for reliable subgoal execution, and termination rules, suitable memory mechanisms, and observation representations provide high-leverage mechanisms for connecting planning with control. Together, these findings point to practical design principles for building more capable and robust hierarchical VLA agents, opening new possibilities for modular embodied systems that combine high-level reasoning with reliable low-level control.

\section{Related Work}

% This study focuses on the black-box scenario, where langauge is the most suitable communication interface. Some other works resort to E.g. points; bounding box, as the interface. Requires specialized data for training both VLA and VLM, not very trivial

% \subsection{Hierarchical VLA Systems}
% \textbf{Hierarchical VLAs} 
While flat VLAs can execute dexterous motion and handle a broad range of tasks, they often struggle with long-horizon and reasoning tasks, due to a lack of coverage of the training data~\cite{black2024pi_0} and the catastrophic forgetting during finetuning~\cite{french1999catastrophic}. The idea of hierarchy \cite{brooks2003robust, sutton1999between, ahn2022can, hu2025slac, su2025hitter, kahneman2011thinking, Fu2026CaPXAF, Shi2025MaestroOR} offers a natural way to decompose such tasks and enable VLAs to achieve compositional generalization.
Hierarchical VLA systems combine the reasoning capabilities of large VLMs with the control fidelity of low-level VLA or skill-conditioned policies. This paradigm has been adopted in many state-of-the-art systems, including G0~\cite{jiang2025galaxeaopenworlddatasetg0}, Humanoid-VLA~\cite{ding2025humanoid}, RoboOS ~\cite{tan2025roboos}, Hi-Robot~\cite{shi2025hi}, Pi-0.5~\cite{intelligence2025pi_}, Gemini Robotics 1.5~\cite{team2025gemini}, HAMSTER~\cite{li2025hamster}, and Helix~\cite{figure2024helix}.
Despite this rapid empirical progress, however, there remains limited understanding of what components matter most for the effectiveness of hierarchical VLA systems. Questions such as how model capabilities, memory handling, or policy switching mechanisms influence task success remain largely unexplored. This paper takes the first step toward a systematic analysis of hierarchical VLA architectures, with
a focus on the system-design layer on top of existing VLM and VLA backbones.
% Through identifying the key design choices and bottlenecks that most strongly impact task performance, we aim to provide architecture-agnostic guidelines that apply across a broad class of hierarchical VLA systems.
We defer detailed discussion of flat VLAs and their evaluations to Appendix~\ref{rw:flat}.

\section{A Unified View of Hierarchical VLAs}

% The rapid developments in hierarchical vision-language-action models present challenges for systematic study and evaluations. To overcome these challenges, 
The first step in our analysis seeks to encapsulate different hierarchical VLA design choices under a shared framework.
In particular, we notice that seemingly different hierarchical systems can all be subsumed by a unified control loop that resembles the options framework~\cite{sutton1999between}. 
In a hierarchical VLA system formed by a \textbf{high-level VLM} and a \textbf{low-level VLA}, the VLA can be viewed as an intra-option policy $\pi_{\textit{VLA}}(a|o, l)$ that maps language instruction $l\in L$ and observation $o \in O$ to low-level robot actions $a \in A$; whereas the high-level VLM can be viewed as an option selection policy $\pi_{\textit{VLM}}(l|\boldsymbol{o}, I)$ that maps (a set of) observations $\boldsymbol{o}$ and a task instruction $I$ to some language instruction $l$. 
Additionally, the observation input to the VLM is managed by a \textbf{memory module} $\boldsymbol{o} \leftarrow \text{mem}([o_i]_{i \leq t})$ that processes historical interactions; as well as an \textbf{observation representation module} $o' \leftarrow \phi(o)$ that optionally post-processes the raw image observations.
Together, the VLM, VLA, memory module and the observation representation module define a hierarchical visuomotor policy over low-level robot actions: 
% $\pi_{\textit{HiVLA}}(a | [o_i]_{i \leq t}, I) = \int_l \pi_{\textit{VLA}}(a | o_t, l) \, \pi_{\textit{VLM}}(l | \boldsymbol{o}, I)\, dl$, where $\quad \boldsymbol{o} = \text{mem}([\phi(o_i)]_{i \leq t}$.
\begin{equation}
\begin{aligned}
\pi_{\textit{HiVLA}}(a \mid [o_i]_{i\le t}, I)
&=
\int_l \pi_{\textit{VLA}}(a \mid o_t,l)\,
\pi_{\textit{VLM}}(l \mid \boldsymbol{o},I)\,dl, \\
&\text{where } \boldsymbol{o}
= \mathrm{mem}([\phi(o_i)]_{i\le t}).
\end{aligned}
\end{equation}

In practical systems, $\pi_{\textit{VLA}} $ and $\pi_{\textit{VLM}}$ typically do not operate at the same frequency. The VLM operates at a much lower frequency due to the high inference cost, where the same language instruction $l$ is 

\begin{minipage}[t]{0.40\linewidth}
\vspace{-12pt}
kept fixed for multiple steps until a \textbf{termination condition} $\beta(o,t)$ is met. Once the termination 
condition is met, the VLM generates a new temporally extended language command $l'$.\\

We present this unified control loop in Alg.~\ref{app:alg}, which \emph{allows us to discuss each component of the hierarchical VLA as a function that can be implemented in different ways}. For example, in previous works, the high-level VLM has been implemented with PaliGemma~\cite{intelligence2025pi_} or finetuned Gemini~\cite{team2025gemini}, while the termination condition has been implemented as a success detector~\cite{team2025gemini} or as a fixed timer~\cite{figure2024helix}.
We base our subsequent experiments on this unified control loop, where we examine the effect of different implementations of each component.
\end{minipage}
\hspace{0.03\linewidth}
\begin{minipage}[t]{0.55\linewidth}
\vspace{-12pt}
\small
% \begin{center}
% \begin{minipage}{0.80\textwidth}
\begin{algorithm}[H]
\DontPrintSemicolon
\caption{Unified Hi-VLA Control Loop
}
\label{app:alg}

% --- Inputs & Outputs ---
\KwIn{Task instruction $I$}

% --- Initialization ---
\BlankLine
\textbf{Initialize:} $t \leftarrow 0$, env $\mathcal{E}$, observation history $\mathbf{o} \leftarrow \emptyset$\;
\textbf{Functions:} Termination $\beta(\cdot)$, Memory $\mathrm{mem}(\cdot)$, \\\hspace*{5em}Obs. Rep. $\phi(\cdot)$, Policies $\pi_{\textsc{vlm}}$ and $\pi_{\textsc{vla}}$\;
\BlankLine

% --- Main Loop ---
\While{not $\mathcal{E}.done$}{
    \tcp*[l]{\colorbox{cyan!15}{High-level VLM Execution Loop}}
    $o_t \leftarrow \mathcal{E}.\mathrm{observe}()$\;
    $o'_t \leftarrow \phi(o_t)$ \hfill \tcp*[r]{\small Process image}
    $\mathbf{o} \leftarrow \mathrm{mem}([o'_i]_{i \leq t})$ \hfill \tcp*[r]{\small Update memory}
    $l \sim \pi_{\textsc{vlm}}(l \mid \mathbf{o}, I)$ \hfill \tcp*[r]{\small VLM inference}
    
    \BlankLine
    \While{not $\beta(o_t, t)$}{
        \tcp*[l]{\colorbox{green!10}{Low-level VLA Execution Loop}}
        $a \sim \pi_{\textsc{vla}}(a \mid o_t, l)$ \hfill \tcp*[r]{\small VLA inference}
        $o_t \leftarrow \mathcal{E}.\mathrm{step}(a)$\tcp*[r]{\small Env Interaction}
        $t \leftarrow t + 1$\;
    }
}
\end{algorithm}
% \end{minipage}
% \end{center}

% \begin{algorithm*}[t!]
% \caption{A Unified Hierarchical VLA Control Loop \jiaheng{make this a bit prettier}}
% \label{app:alg}
% \textbf{Initialize} environment $\mathit{env}$ , task instruction $I$,  total time horizon $T$, and current time $t \leftarrow 0$\;
% \textbf{Load} Low-level VLA policy $\pi_{\textit{VLA}}(a | o_t, l)$, and High-level VLM policy $\pi_{\textit{VLM}}(l | \boldsymbol{o}, I)$ \;
% \textbf{Initialize} termination function $\beta(o)$, memory function $mem([o])$, and observation representation $obs\_rep(o)$  \;
% \While{not $env.done$}{
%     $o_t \leftarrow \mathit{env.observe}()$ \;
%     $o'_t \leftarrow \mathit{obs\_rep}(o_t)$ \tcp*{Process observations}
%     $\boldsymbol{o} \leftarrow \mathit{mem}([o'_i]_{i \leq t}) $  \;
%     $l \sim \pi_{\textit{VLM}}(l | \boldsymbol{o}, I)$  \tcp*{Step high-level VLM}
%     \While{not $\beta(o_t)$}{
%         $a \sim \pi_{\textit{VLA}}(a | o_t, l)$\tcp*{Step the VLA until termination condition is met}
%         $o_t \leftarrow \mathit{env.step}(a)$ \; 
%         $t \leftarrow t + 1$ \; 
%         \If{$t \geq T$
%         }{Terminate\tcp*{Finish the episode if reach time limit}}
%     }
% }
% \end{algorithm*}
\end{minipage}

%%%%%%%%%%%%%%%%%%%%%%%%%%%%%%%%%%%%%%%

%%%%%%%%%%%%%%%%%%%%%%%%%%%%%%%%%

\section{A Systematic Study of Hierarchical VLA Agents}

\subsection{Evaluation Setup}

We conduct our main experiments in the MuJoCo ALOHA suite: a simulated table-top manipulation benchmark with demonstrated real-to-sim transferability~\cite{team2025gemini}. 
Additionally, we run experiments on a real ALOHA robot. 
We categorize our evaluation tasks into three categories, namely \textbf{short-horizon, long-horizon,} and \textbf{reasoning} tasks. We discuss our detailed experimental setup in Appendix~\ref{sup:eval}.

\begin{figure*}[t!]
    \vspace{-0.2cm}
    \centering
    \begin{subfigure}[t]{0.99\textwidth}
        \includegraphics[width=\textwidth]{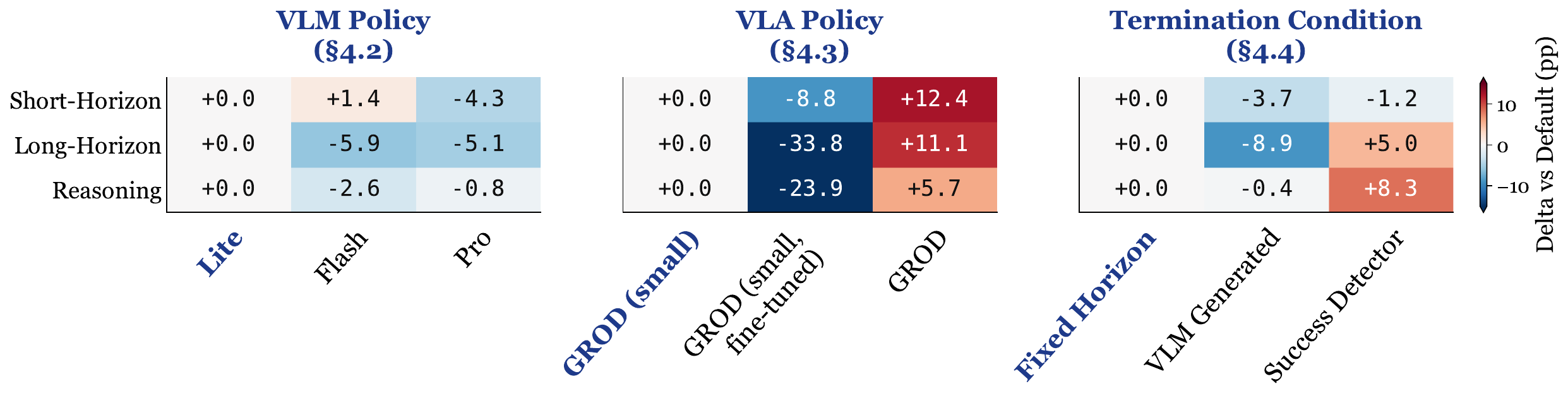}
    \end{subfigure}
    \begin{subfigure}[t]{0.99\textwidth}
        \includegraphics[width=\textwidth]{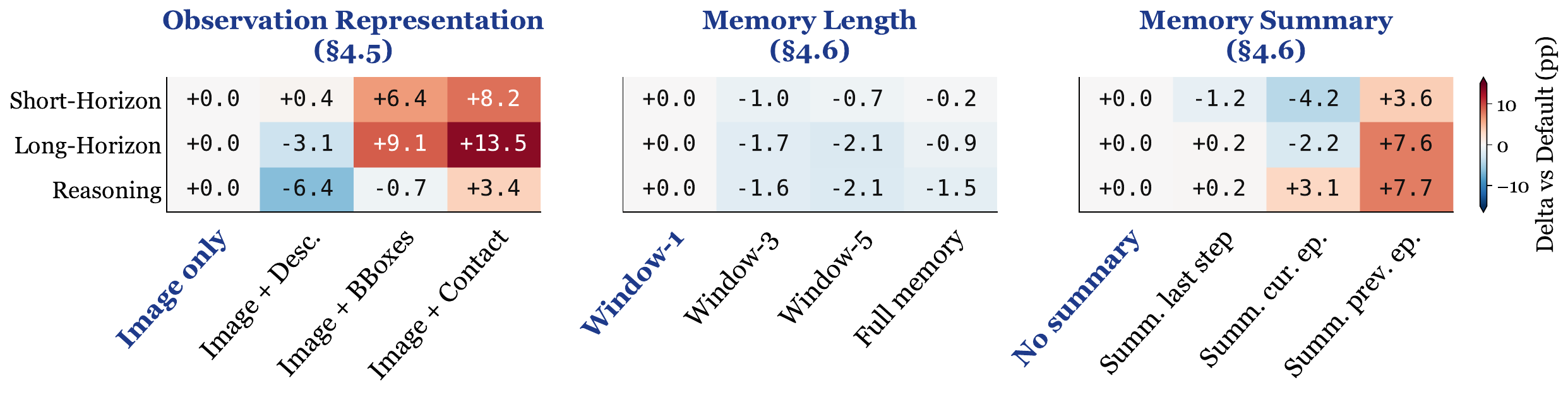}
    \end{subfigure}
    \vspace{-0.3cm}
    \caption{\small An overview of our experimental results. In these plots, we visualize how different design choices increase (red) or decrease (blue) the overall performance of hierarchical VLA systems on different types of tasks. We show the detailed results in Appendix~\ref{sup:results}.}
    \label{fig:results}
    \vspace{-0.5cm}
\end{figure*}

In the following sections, we present the results and analysis for a comprehensive set of controlled experiments designed to systematically evaluate the effect of different hierarchical VLA components presented in Alg.~\ref{app:alg}, including the high-level VLM policy (Sec.~\ref{ss:vlm}), the low-level VLA policy (Sec.~\ref{ss:vla}), the termination condition (Sec.~\ref{ss:term}), the observation representation (Sec.~\ref{ss:obs}), and the memory system (Sec.~\ref{ss:mem}). 
For each component study, we vary one component while holding the rest of the system fixed, which allows us to attribute performance changes to the component being studied rather than unrelated stack differences. Finally, we show that these design choices collectively lead to improved performance over both flat VLA and a naive hierarchical VLA (Sec.~\ref{ss:agg}), and demonstrate that our conclusions extend beyond our primary evaluation stack and generalize to a different VLM, VLA, benchmark, and robot embodiment (Sec.~\ref{ss:cross_stack}).

\subsection{Effect of High-level VLM Policy}
\label{ss:vlm}

The high-level VLM policy is responsible for aggregating information and generating language commands for the low-level VLA to execute (prompt in Appendix~\ref{sup:vlm_prompt}).
Our first set of experiments evaluates the impact of the VLM on the eventual performance.
% , with a focus on the impact of model size and reasoning capabilities.
%%%%%%%%%%%%%%%%%%%%%%%%%%%%%%%%%
%%%%%%%%%%%%%%%%%%%%%%%%%%%%%%%%%

\begin{wrapfigure}{r}{0.35\textwidth}
  \vspace{-0.35cm}
  \centering
  \includegraphics[width=0.35\textwidth]{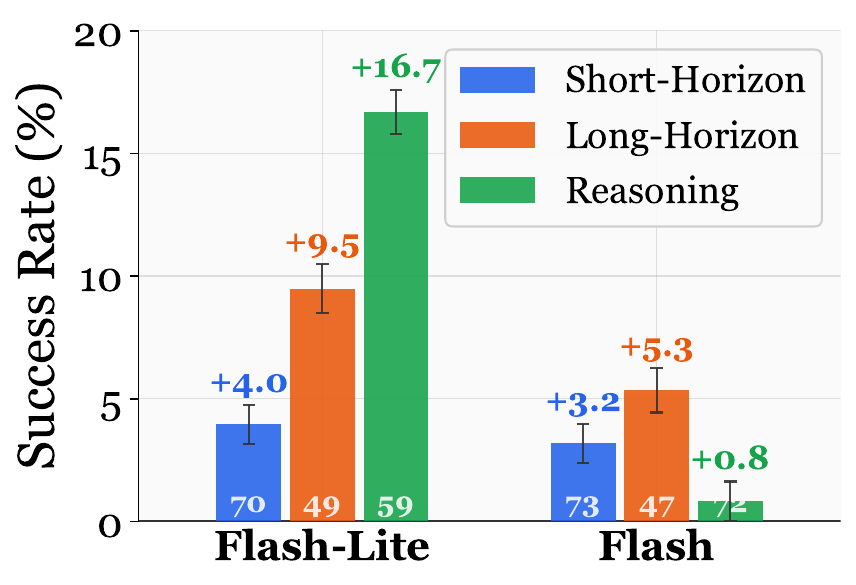}
  \caption{\small Change in success rates after \textbf{adding VLM thinking}.}
  \label{fig:wrapped}
  \vspace{-0.3cm}
\end{wrapfigure}

To vary VLM model size and reasoning capability while holding other factors constant, we focus on a single model family, Gemini, which allows us to minimize confounding differences in model lineage, training data, and interfaces. Specifically, we evaluate models from the Gemini 2.5 series, spanning a range of capabilities from the resource-efficient Lite model, through Flash, to the more capable Pro model~\cite{comanici2025gemini}. For Lite and Flash, we evaluate two different inference modes: with and without the \textit{thinking} capability toggled on. When the thinking is toggled on, the model runs multiple internal passes that generate, criticize, and refine the output, which enhances its reasoning capabilities at the cost of slightly lower inference speed. We evaluate 2.5 Pro only with thinking on, since the Pro does not allow disabling the thinking capabilities. We visualize the results in Fig.~\ref{fig:results} top-left and Fig.~\ref{fig:wrapped}, and report the detailed results in Table~\ref{tab:vlm}.

Across all task categories and models, VLM inference with thinking enabled consistently outperforms the counterpart (Fig.~\ref{fig:wrapped}), indicating that the reasoning capability of the VLM is critical towards better performance of hierarchical VLA. This is likely due to the fact that effectively making use of all the information available to the high-level policy is non-trivial, and thinking allows the model to better utilize this information (see Fig.~\ref{fig:pull} for an example). Additionally, notice that long-horizon tasks benefit more from thinking compared to the short-horizon tasks, suggesting that the reasoning capability becomes more important as the task gets more complex.

Surprisingly, the  size of the VLM does not have a significant impact on performance, where Lite, Flash, and Pro have similar performance across the board when thinking is on (Fig.~\ref{fig:results} top-left). This is consistent with how existing hierarchical systems such as Pi-0.5 and Pi-0.7~\cite{intelligence2025pi_,intelligence2026pi07steerablegeneralistrobotic} can be performant despite using smaller high-level VLMs. 
One possible interpretation is that most existing robotics benchmark tasks do not require knowledge and instruction following ability beyond what smaller VLMs such as Lite can already provide, although larger models may become more important for tasks involving unfamiliar interfaces (e.g. operating a new coffee machine). 
Thus, despite Gemini-Pro outperforming Flash and Lite on existing VLM benchmarks~\cite{comanici2025gemini}, our results suggest that such benchmarks may not directly predict performance in hierarchical VLA systems.

\begin{summary}%[Key Takeaways]
\textbf{Key Takeaways.} Improved reasoning capability (via thinking) of the VLM has a big impact on the overall performance of the system. By comparison, the model size of the VLM seems to matter much less, where smaller reasoning-based models can work as well as larger models. 
% \begin{itemize}
%     \item Improved reasoning capability (via thinking) of the VLM has a big impact on the overall performance of the system.
%     \item By comparison, the model size of the VLM seem to matter much less, where smaller reasoning-based models can work as well as larger models.
% \end{itemize}
\vspace{-2pt}
\end{summary}

\subsection{Effect of Low-Level VLA Policy}

\label{ss:vla}

Next, we examine the impact of different low-level VLAs on the overall performance of the hierarchical system, where we focus on evaluating the impact of the size of the VLA as well as the effect of fine-tuning. We use a family of Gemini Robotics On-Device (GROD) Model~\cite{team2025gemini} for our experiments, which again allows us to control and avoid differences in model lineage, training data, and interfaces. Specifically, we test three types of VLAs: GROD model trains with only real robot data; GROD (small), a smaller model trained with the same real dataset; and GROD (small) model, which is post-trained with in-domain simulation demonstrations from the same Mujoco environment that we evaluated in. We visualize the results in Fig.~\ref{fig:results} top-mid, and report the detailed results in Table~\ref{tab:vla}.

We can see that changing the VLA has a big effect on the performance of the system. This is unsurprising considering that the low-level VLA is ultimately the module that controls the robot. More specifically, we noticed that the GROD model with larger number of parameters consistently shows better performance than the smaller model, due to its better instruction following (as shown by the strong performance on short-horizon tasks) and generalization capability. Interestingly, the smaller GROD model finetuned with in-domain simulation data gives the worst performance, especially for long-horizon tasks. This is likely due to the fact the fine-tuning often results in worse instruction following capability of the VLA~\cite{Huang2026BreakingLP,Chen2026SteerableVP} (i.e., slight rephrasings of the same instruction), which turns out to be very critical for maintaining good hierarchical performance.

\begin{summary}
\textbf{Key Takeaways.} Unlike high-level VLMs, low-level VLAs benefit significantly from increased size, likely due to the improved instruction following capabilities, making it more suitable for VLM orchestration.
Simultaneously, loss of VLA steerability can lead to significant drop in performance, as shown by the poor performance of the simulation fine-tuned GROD model.
% \begin{itemize}
%     \item Unlike VLMs, low-level VLA benefit significantly from increased size, likely due to the improved instruction following capabilities, making it more suitable for VLM orchestration.
%     \item Simultaneously, loss of VLA steerability can lead to significant drop in performance, as shown by the poor performance of the simulation fine-tuned VLA. 
% \end{itemize}
\vspace{-2pt}
\end{summary}

\subsection{Termination Conditions}
\label{ss:term}
Similar to the options framework, a critical design choice in Hi-VLAs is the choice of when to hand the control back to the high-level VLM, also known as the ``termination condition.'' We evaluate three types of termination conditions in our work: 

\begin{itemize}
    \item \textbf{Fixed Frequency:} The VLA hands control back to the VLM at a fixed, preset frequency.
    \item \textbf{Success Detection:} A success detector decides whether the instruction generated by the VLM is successfully completed given the current state. % Control is handed back to the VLM whenever success is detected.
    \item \textbf{VLM Termination:} The VLM generates an ``expected execution time'' along with the language command whenever it is queried. % The VLA hands control back to the VLM when the expected time is reached.
\end{itemize}

In this work, we implement the success detection by querying a VLM with privileged state of the simulator to make it as accurate as possible, where the prompt is shown in Appendix~\ref{sup:succ_det}.
We visualize the results in Fig.~\ref{fig:results} top-right, and report the detailed results in Table.~\ref{tab:term}.
Among the three tested methods, the success detector consistently achieves good performance, showing that having a good termination condition can indeed positively impact the performance. VLM termination performs the worst, likely due to the stochastic nature of the low-level VLA which makes it hard to accurately predict execution length in advance. Finally, the short-horizon tasks are relatively agnostic to the termination conditions, likely because they do not require command sequencing and is thus less affected by when to switch.

\begin{wrapfigure}{r}{0.33\textwidth}
  \vspace{-0.35cm}
  \centering
        \includegraphics[width=0.33\textwidth]{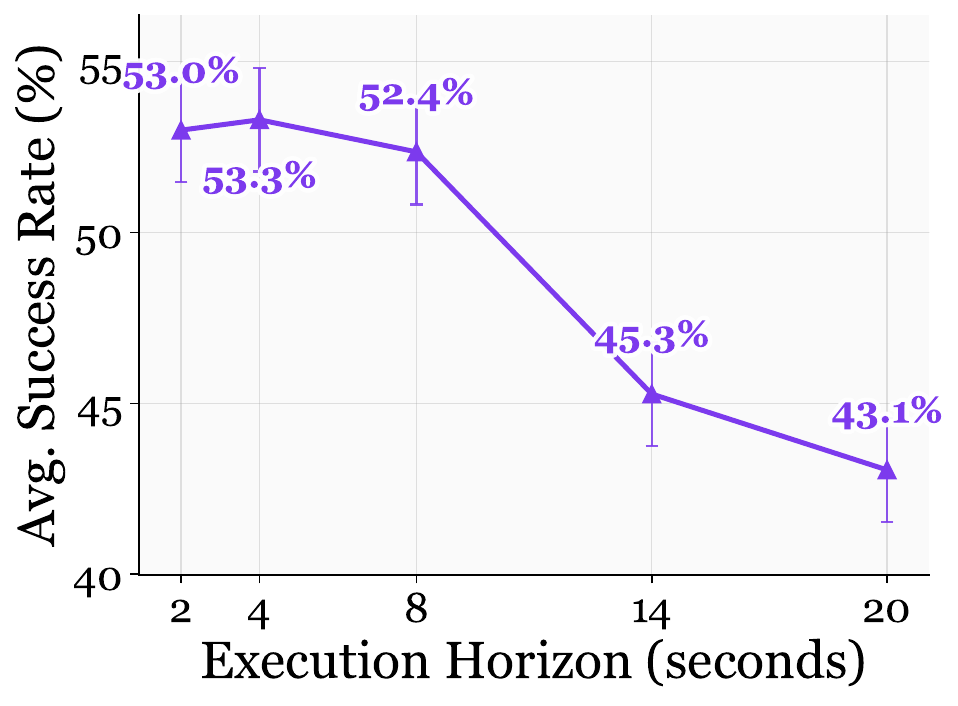}
        \caption{\small VLA Exec. Horizon}
        \label{fig:temp}
      \centering
      \vspace{0.55cm}
        \includegraphics[width=0.33\textwidth]{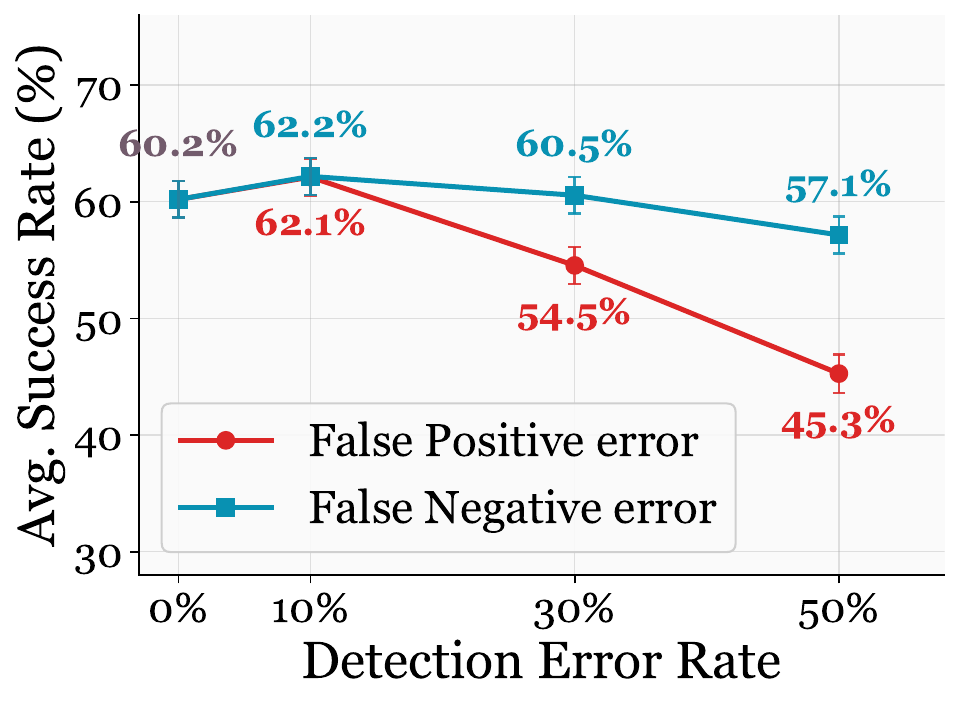}
        \caption{\small Success Detection Error}
        \label{fig:succ_fp}
  \vspace{-0.3cm}
\end{wrapfigure}

\textbf{What is the effect of the VLA execution horizon?} An important hyperparameter for Fixed Frequency Termination is the \emph{execution horizon}, which controls the number of low-level VLA steps before handing control back to the VLM. We conduct additional experiments evaluating this important hyper-parameter, and present the results in Fig.~\ref{fig:temp}.
As shown by the results, a too-long horizon can lead to timeouts for multi-step tasks, causing a significant drop in performance.
% (the VLM planner can compensate for short execitopm by asking the VLA to execute the same instruction again if needed)
While shorter execution horizon generally improves performance, the computational cost of higher-frequency VLM query makes it less ideal. Overall, we recommend selecting a moderate horizon, e.g., 4-8 seconds, that reduces the computational costs of VLM queries while maintaining a frequent-enough VLM control shown by very little performance drop.

% Interesting, once the temporal extension parameter is below a certain threshold, having faster switching no longer affects the success rate much, but starts to incur significantly more computational cost due to the costly VLM queries. 

\textbf{What if the success detector is not accurate?} While having a success detector as a termination condition can improve performance, obtaining a good success detector is not always easy~\cite{du2023vision}. 
Therefore, a natural question is how the system will behave as the accuracy of the success detector deteriorates. Our results (Fig.~\ref{fig:succ_fp}) show that success detector is a robust termination condition under moderate error. We elaborate on this experiment in Appendix~\ref{sup:succ_rob}. 

\begin{summary}
\textbf{Key Takeaways.} 
Success detection, even with moderate detection error, can be a powerful termination condition. However, too high of a detection error will significantly impact its performance.
For fixed-horizon termination, a good VLA execution horizon (e.g. 4-8 seconds) allows us to reduce the cost of frequent VLM calls without significantly sacrificing the performance.
% \begin{itemize}
%     \item For fixed-horizon termination, a good VLA execution horizon (e.g. 4-8 seconds) allows us to reduce the cost of frequent VLM calls without sacrificing the performance of the system.
%     \item Success detection, even with moderate detection error, can be a powerful termination condition. However, too high of a detection error will significantly impact its performance.
%     % \item Compared to false-negative error, false-positive errors are more harmful to the performance of the system.
% \end{itemize}
\vspace{-2pt}
\end{summary}

\subsection{Observation Representations}
\label{ss:obs}

Whenever the high-level VLM makes decisions, it needs to know the current state of the robot. While the naive approach would be to rely entirely on the image observation, we found that we often see better performance by carefully processing these image observations into text descriptions. Here, we examine four types of observation representation methods:
\begin{itemize}
    \item \textbf{Raw Image}: Where we do not pass in any additional text description.
    \item \textbf{Naive Summarization}: Where we ask the VLM to first describe the image.
    \item \textbf{Summarize with Privileged Info}: Where we additionally pass in privileged information (more specifically, contact between objects) from the simulator when generating the text description.
    \item \textbf{Summarize with Bounding Box}: Where we first ask the VLM to generate bounding boxes of the objects, and use that information to generate the text description.
\end{itemize}

We visualize these representations in Fig.~\ref{fig:obs_rep}, and show the prompt for text summarization and bounding box generation in Appendix~\ref{sup:obs_desc_prompt} and Appendix~\ref{sup:bb_desc_prompt}. We visualize the results in Fig.~\ref{fig:results} bottom-left, and present the detailed results in Table~\ref{tab:obs_representations}.
Results show that both adding bounding box information and adding privileged information from simulator when generating text description significantly boost performance. This result is quite interesting since ideally the raw image already contains all the information, and we shouldn't have to pass in additional text description. One explanation for this could be due to the phenomenon that VLMs tend to ignore image inputs as task becomes harder~\cite{majumdar2024openeqa}, which is why passing in additional text information will boost performance.

Given that privileged information gives us the best performance across the board, we anticipate future enhancements in the image processing and \emph{spatial understanding} capabilities of the VLM and/or incorporation of extra sensor modality such as tactile sensors to be important to the performance of hierarchical VLAs.

%%%%%%%%%%%%%%%%%%%%%%%%%%%%%%%%%%%%%%%

\begin{summary}
\textbf{Key Takeaways.} 
Good observation representation is critical to the performance of Hi-VLA. For example, bounding box description notably boost performance without requiring any extra information.
Adding privileged information can further boost performance, calling for more focus on improving spatial understanding capabilities of the VLM.
% \begin{itemize}
%     \item Good observation representation is critical to the performance of Hi-VLA. For example, bounding box description notably boost performance without requiring any extra information.
%     \item Adding privileged information can further boost performance, calling for more focus on improving spatial understanding capabilities of the VLM.
% \end{itemize}
\vspace{-2pt}
\end{summary}

\begin{figure*}[t]
    \vspace{-0.2cm}
    \centering
    \includegraphics[width=0.99\linewidth]{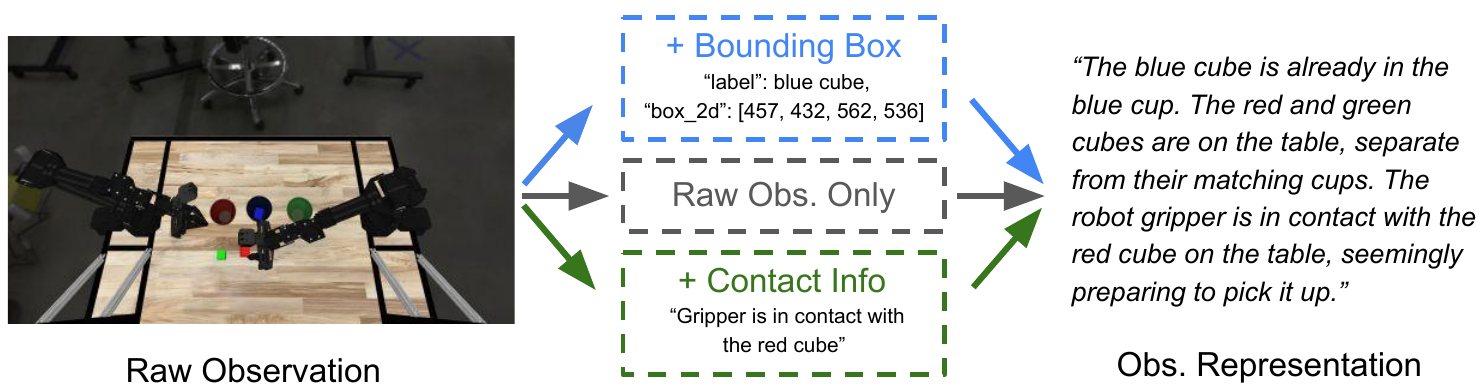}
    \caption{\small Observation representation pipeline. We study three ways of converting the raw image observation to text: (1) querying a VLM naively, (2) incorporating (VLM-generated) bounding box information to the query, and (3) incorporating privileged contact information to the query.}
    \label{fig:obs_rep}
    \vspace{-0.5cm}
\end{figure*}

\subsection{Memory}
\label{ss:mem}

In this section, we examine the effect of appending previous interactions to the VLM context. 
% Notice that the purpose here is not to test the ability of tackling partially observable tasks (since all of our tasks are fully observable), but rather to test whether the VLM can benefit from autonomous in-context exploration. 
Specifically, we accumulate the VLM history for 1 step, for 3 steps, for 5 steps, and from the entire episode so far, and feed it to the VLM planner. 
We visualize the results in Fig.~\ref{fig:results} bottom-mid, and present the detailed results in Table~\ref{tab:mem}.
We find that the length of history context doesn’t affect the performance much. This result suggests that the VLM cannot extract useful information from raw history of current episode, possibly because there is too little information to learn from.

%%%%%%%%%%%%%%%%%%%%%%%%%%%%%%%%%%%%%%%%%

\textbf{Experience summarization and reflection.}
Memory does not necessarily have to appear in raw form. Instead, we can distill useful information from raw memory via summarization and reflection~\cite{park2023generative,shinn2023reflexion}. We test summarizing experiences across 1-step, the current episode, and from previous episodes. Specifically for the previous episode setup, we first roll out the system for 10 episodes of the same task, and then let a VLM summarize the experiences into affordances. We show the prompt for memory summarization in Appendix~\ref{sup:mem_prompt}.
We visualize the results in Fig.~\ref{fig:results} bottom-right, and present the detailed results in Table~\ref{tab:mem_sum}.

We find that in-episode summarization generally has a neutral to negative effect on performance, as seen when comparing “Memory Window - 1” with “Summary of last step” and “Full Memory” with “Summary of current episode.” However, summarizing experiences across episodes can positively impact performance. This suggests that for hierarchical systems, extracting affordances from cross-episode information (especially previous successful episodes) is more beneficial than relying on in-episode failure signals for on-the-fly VLM corrections. Future work could explore more powerful techniques, such as reinforcement learning or supervised finetuning of the VLM, to better leverage cross-episodic interactions with the VLA.

%  Upon examining the episodes, we noticed that
% If the current episode is exploring in the right direction, we can usually solve the task without in-context reflection
% If it is exploring in the wrong direction, it is hard to use that information effectively, since knowing that some action does not work provides quite limited information

\begin{summary}
\textbf{Key Takeaways.} For fully observable table-top tasks, the current system does not benefit much from in-episode memory even when summarization is available, motivating more effective memory processing and in-context information extraction capabilities. However, cross-episodic knowledge helps a lot with task completion. A fruitful future direction is to explore how to perform VLM post-training with these cross-episodic experiences.
% \begin{itemize}
%     \item The current system does not benefit much from in-episode memory even when summarization is available, and lacks the ability to do in-context learning, calling for more sophisticated memory processing mechanism. 
%     \item However, cross-episodic knowledge helps a lot with task completion. A fruitful future direction may be to explore how to perform VLM post-training with these cross-episodic experiences.
% \end{itemize}
\vspace{-2pt}
\end{summary}

\subsection{Aggregation of Discoveries}
\label{ss:agg}

To put all our discoveries together, we evaluate:
\begin{itemize}
    \item \textbf{Best hierarchical system}: the system that takes the combination of best-performing parameters from each of the earlier experiments, i.e. cross-episodic memory, thinking VLM, contact-based observation description, and success-based termination condition. 
    \item \textbf{Naive hierarchical system}: the system that takes the combination of naively-chosen parameters from earlier experiments, i.e. no memory, VLM without thinking, raw observation, and fixed-horizon termination.
    \item \textbf{Flat VLA}: No VLM planner. The task prompt is directly passed to the low-level VLA.
\end{itemize}

Importantly, for each evaluation task, all three setups have the same VLA and input task prompt. We present the results in Table~\ref{tab:agg}.
We can see that even a naive implementation of a hierarchical system often outperforms the flat architecture, demonstrating the importance of introducing hierarchy. However, as the task becomes more challenging, the naive implementation quickly fall short, where more sophisticated hierarchical systems lead to a greater performance uplift. 
% \subsection{Real Robot}
% \label{ss:real}
We further test these conclusions on a \textbf{real robot}, where we command an ALOHA robot to place fruits onto plates of matching color (Fig.~\ref{fig:motion_sequence}). Unlike the simulator Best Hierarchy, the real-robot configuration uses image-derived bounding-box observations and a success detector that receives no contact or simulator-state information. Across 5 trials, it correctly places 12/15 fruits, compared with 9/15 for the Naive Hierarchy and 3/15 for the flat VLA (Table~\ref{tab:agg}, right), demonstrating that the practical, non-privileged hierarchy retains a substantial advantage. Lastly, we experiment on how potential improvements in VLA capabilities may affect our conclusions, and discuss the results in Appendix~\ref{sup:future}.

\begin{table*}[t!]\centering
% Set up alternating row colors starting from the second row (the data rows)
% \rowcolors{2}{white}{lightgray}
\setlength{\tabcolsep}{8pt} % Increase space between columns
\footnotesize
\begin{tabular}{l cccc}
\toprule
\textbf{Configuration} & \textbf{Short-Horizon} ($\%$) & \textbf{Long-Horizon} ($\%$) & \textbf{Reasoning} ($\%$) & \textbf{Real ALOHA}\\
\midrule
Best Hierarchy & \textbf{78.22 $\pm$ 0.91} &  \textbf{67.08 $\pm$ 1.38} & \textbf{80.89 $\pm$ 1.17} & \textbf{12 / 15}\\
Naive Hierarchy & 69.57 $\pm$ 1.15 & 40.56 $\pm$ 1.37 & 66.49 $\pm$ 1.31 & 9 / 15\\
Flat VLA & 69.63 $\pm$ 1.07 & 25.30 $\pm$ 1.22 & 50.90 $\pm$ 1.20 & 3 / 15 \\
\bottomrule
\end{tabular}
\caption{\small Performance of the best hierarchy, naive hierarchy, and flat VLA.}
\vspace{-0.3cm}
\label{tab:agg}
\end{table*}

\begin{summary}
\textbf{Key Takeaways.} 
A system with a clear, hierarchical control structure (orchestration) significantly boosts performance compared to a flat structure. However, simply introducing hierarchy is not enough; a good implementation of orchestration can make a big difference, especially for long-horizon and reasoning tasks, and will remain as VLA capabilities improve in the future.
% \begin{itemize}
%     \item Orchestration is crucial: A system with a clear, hierarchical control structure (orchestration) significantly boosts performance compared to a flat structure.
%     \item Quality of implementation matters: simply introducing hierarchy is not enough; a good implementation of orchestration can make a big difference, especially for long-horizon and reasoning tasks, and will remain as VLA capabilities get better in the future.
%     % \item There is significant room for improvement. Future works can further enhance the capabilities of these hierarchical agents by exploring the directions outlined in previous sections.
% \end{itemize}
\vspace{-2pt}
\end{summary}

% SUGGESTED ADDITION START: cross-stack validation and steerability analysis.
% Accept by removing the \chadded wrappers (or loading changes with the final
% option); reject by deleting this entire marked block.
\subsection{Cross-Stack Validation}
\label{ss:cross_stack}

To test whether our conclusions transfer to a different stack, we combine Qwen3-VL planners~\cite{bai2025qwen3vl}, $\pi$-family VLAs~\cite{intelligence2025pi_}, and LIBERO~\cite{liu2023libero}. We evaluate three long-horizon, reasoning tasks adapted from LIBERO-Object, with 150 trials per task. We elaborate on these tasks in Appendix~\ref{sup:cross_stack_tasks}.

\begin{table}[H]
\centering
\footnotesize
\setlength{\tabcolsep}{4pt}
\begin{tabular*}{0.98\linewidth}{@{\extracolsep{\fill}}l c l c l c@{}}
\toprule
\multicolumn{2}{c}{\textbf{Hierarchy}} &
\multicolumn{2}{c}{\textbf{VLM Scale}} &
\multicolumn{2}{c}{\textbf{VLA Steerability}} \\
\cmidrule(lr){1-2}\cmidrule(lr){3-4}\cmidrule(lr){5-6}
\textbf{Configuration} & \textbf{Success (\%)} &
\textbf{Configuration} & \textbf{Success (\%)} &
\textbf{Configuration} & \textbf{Success (\%)} \\
\midrule
Best hierarchy & $83.8 \pm 3.0$ &
Qwen3-VL 4B & $76.0 \pm 3.5$ &
$\pi_{0.5}$ & $73.3 \pm 3.6$ \\
Naive hierarchy & $70.0 \pm 3.7$ &
Qwen3-VL 8B & $70.7 \pm 3.7$ &
Narrow-FT $\pi_{0.5}$ & $32.0 \pm 3.8$ \\
Flat VLA & $43.3 \pm 4.0$ &
Qwen3-VL 32B & $73.3 \pm 3.6$ &
& \\
\bottomrule
\end{tabular*}
\vspace{0.3cm}
\caption{Cross-stack validation with Qwen3-VL planners, $\pi$-family VLA policies, and LIBERO reasoning tasks. Values report success rates $\pm$ standard errors.}
\label{tab:cross_stack}
\vspace{-0.3cm}
\end{table}

As shown in Table~\ref{tab:cross_stack}, \textbf{our findings continue to hold in this new setting}: increasing VLM size by $8\times$ provides no monotonic improvement (Sec.~\ref{ss:vlm}), while the best hierarchy outperforms both naive hierarchy and flat control (Sec.~\ref{ss:agg}). We also test VLA steerability by narrow-finetuning $\pi_{0.5}$ on a single instruction. The policy overfits to that instruction and becomes less responsive to alternatives, substantially reducing hierarchical performance and supporting our conclusion in Sec.~\ref{ss:vla}.
% SUGGESTED ADDITION END.

\section{Conclusion}
In this work, we presented a systematic, controlled evaluation of design choices within Hierarchical Vision-Language-Action (Hi-VLA) systems, addressing the critical question of "what matters" for performance in complex robotic manipulation tasks. By constructing a flexible framework, we rigorously benchmarked the impact of various high-level VLMs, low-level VLAs, termination conditions, observation modalities, and memory mechanisms across diverse task genres including short-horizon, long-horizon, and reasoning challenges. Our key findings provide concrete guidance for researchers and practitioners to design future Hi-VLA systems.

\textbf{Limitations and Future Work.}
A limitation of our current study is its focus on static environments and the assumption that latency is not a critical performance factor. 
Future work should explicitly investigate the impact of latency-sensitive scenarios and dynamic task environments on Hi-VLA performance. Furthermore, promising avenues include exploring Reinforcement Learning or supervised finetuning of the high-level VLM to better integrate cross-episodic knowledge and thereby more effectively guide the VLA's understanding of its low-level capabilities. Last but not least, the hierarchical architecture itself could be leveraged to inform and guide low-level continual policy improvement~\cite{hu2026simple}, leading to more robust and adaptable agents.

\clearpage
% The acknowledgments are automatically included only in the final and preprint versions of the paper.
\acknowledgments{
We thank Travers Rhodes and Kevin Sayed for help on real robot experiments.
We thank Laura Graesser for feedback on the paper, and Yilun Du, Wentao Yuan, Fei Xia, Wenhao Yu, Ted Xiao, Sandy Huang, Martin Riedmiller, and Jinyu Xie for the valuable discussions.}

%===============================================================================

\bibliography{ref}  % .bib

\begin{thebibliography}{52}
\providecommand{\natexlab}[1]{#1}
\providecommand{\url}[1]{\texttt{#1}}
\expandafter\ifx\csname urlstyle\endcsname\relax
  \providecommand{\doi}[1]{doi: #1}\else
  \providecommand{\doi}{doi: \begingroup \urlstyle{rm}\Url}\fi

\bibitem[Team et~al.(2025)Team, Abeyruwan, Ainslie, Alayrac, Arenas, Armstrong,
  Balakrishna, Baruch, Bauza, Blokzijl, et~al.]{team2025gemini}
G.~R. Team, S.~Abeyruwan, J.~Ainslie, J.-B. Alayrac, M.~G. Arenas,
  T.~Armstrong, A.~Balakrishna, R.~Baruch, M.~Bauza, M.~Blokzijl, et~al.
\newblock Gemini robotics: Bringing ai into the physical world.
\newblock \emph{arXiv preprint arXiv:2503.20020}, 2025.

\bibitem[Black et~al.(2024)Black, Brown, Driess, Esmail, Equi, Finn, Fusai,
  Groom, Hausman, Ichter, et~al.]{black2024pi_0}
K.~Black, N.~Brown, D.~Driess, A.~Esmail, M.~Equi, C.~Finn, N.~Fusai, L.~Groom,
  K.~Hausman, B.~Ichter, et~al.
\newblock $pi\_0 $: A vision-language-action flow model for general robot
  control.
\newblock \emph{arXiv preprint arXiv:2410.24164}, 2024.

\bibitem[Kim et~al.(2024)Kim, Pertsch, Karamcheti, Xiao, Balakrishna, Nair,
  Rafailov, Foster, Lam, Sanketi, et~al.]{kim2024openvla}
M.~J. Kim, K.~Pertsch, S.~Karamcheti, T.~Xiao, A.~Balakrishna, S.~Nair,
  R.~Rafailov, E.~Foster, G.~Lam, P.~Sanketi, et~al.
\newblock Openvla: An open-source vision-language-action model.
\newblock \emph{arXiv preprint arXiv:2406.09246}, 2024.

\bibitem[Zitkovich et~al.(2023)Zitkovich, Yu, Xu, Xu, Xiao, Xia, Wu, Wohlhart,
  Welker, Wahid, et~al.]{zitkovich2023rt}
B.~Zitkovich, T.~Yu, S.~Xu, P.~Xu, T.~Xiao, F.~Xia, J.~Wu, P.~Wohlhart,
  S.~Welker, A.~Wahid, et~al.
\newblock Rt-2: Vision-language-action models transfer web knowledge to robotic
  control.
\newblock In \emph{Conference on Robot Learning}, pages 2165--2183. PMLR, 2023.

\bibitem[Wen et~al.(2025)Wen, Zhu, Li, Zhu, Tang, Wu, Xu, Liu, Cheng, Shen,
  et~al.]{wen2025tinyvla}
J.~Wen, Y.~Zhu, J.~Li, M.~Zhu, Z.~Tang, K.~Wu, Z.~Xu, N.~Liu, R.~Cheng,
  C.~Shen, et~al.
\newblock Tinyvla: Towards fast, data-efficient vision-language-action models
  for robotic manipulation.
\newblock \emph{IEEE Robotics and Automation Letters}, 2025.

\bibitem[Abdolmaleki et~al.(2025)Abdolmaleki, Abeyruwan, Ainslie, Alayrac,
  Arenas, Balakrishna, Batchelor, Bewley, Bingham, Bloesch,
  et~al.]{abdolmaleki2025gemini}
A.~Abdolmaleki, S.~Abeyruwan, J.~Ainslie, J.-B. Alayrac, M.~G. Arenas,
  A.~Balakrishna, N.~Batchelor, A.~Bewley, J.~Bingham, M.~Bloesch, et~al.
\newblock Gemini robotics 1.5: Pushing the frontier of generalist robots with
  advanced embodied reasoning, thinking, and motion transfer.
\newblock \emph{arXiv preprint arXiv:2510.03342}, 2025.

\bibitem[O’Neill et~al.(2024)O’Neill, Rehman, Maddukuri, Gupta, Padalkar,
  Lee, Pooley, Gupta, Mandlekar, Jain, et~al.]{o2024open}
A.~O’Neill, A.~Rehman, A.~Maddukuri, A.~Gupta, A.~Padalkar, A.~Lee,
  A.~Pooley, A.~Gupta, A.~Mandlekar, A.~Jain, et~al.
\newblock Open x-embodiment: Robotic learning datasets and rt-x models: Open
  x-embodiment collaboration 0.
\newblock In \emph{2024 IEEE International Conference on Robotics and
  Automation (ICRA)}, pages 6892--6903. IEEE, 2024.

\bibitem[NVIDIA et~al.(2025)NVIDIA, :, Bjorck, Castañeda, Cherniadev, Da,
  Ding, Fan, Fang, Fox, Hu, Huang, Jang, Jiang, Kautz, Kundalia, Lao, Li, Lin,
  Lin, Liu, Llontop, Magne, Mandlekar, Narayan, Nasiriany, Reed, Tan, Wang,
  Wang, Wang, Wang, Xiang, Xie, Xu, Xu, Ye, Yu, Zhang, Zhang, Zhao, Zheng, and
  Zhu]{nvidia2025gr00tn1openfoundation}
NVIDIA, :, J.~Bjorck, F.~Castañeda, N.~Cherniadev, X.~Da, R.~Ding, L.~J. Fan,
  Y.~Fang, D.~Fox, F.~Hu, S.~Huang, J.~Jang, Z.~Jiang, J.~Kautz, K.~Kundalia,
  L.~Lao, Z.~Li, Z.~Lin, K.~Lin, G.~Liu, E.~Llontop, L.~Magne, A.~Mandlekar,
  A.~Narayan, S.~Nasiriany, S.~Reed, Y.~L. Tan, G.~Wang, Z.~Wang, J.~Wang,
  Q.~Wang, J.~Xiang, Y.~Xie, Y.~Xu, Z.~Xu, S.~Ye, Z.~Yu, A.~Zhang, H.~Zhang,
  Y.~Zhao, R.~Zheng, and Y.~Zhu.
\newblock Gr00t n1: An open foundation model for generalist humanoid robots,
  2025.
\newblock URL \url{https://arxiv.org/abs/2503.14734}.

\bibitem[Belkhale et~al.(2024)Belkhale, Ding, Xiao, Sermanet, Vuong, Tompson,
  Chebotar, Dwibedi, and Sadigh]{belkhale2024rt}
S.~Belkhale, T.~Ding, T.~Xiao, P.~Sermanet, Q.~Vuong, J.~Tompson, Y.~Chebotar,
  D.~Dwibedi, and D.~Sadigh.
\newblock Rt-h: Action hierarchies using language.
\newblock \emph{arXiv preprint arXiv:2403.01823}, 2024.

\bibitem[Intelligence et~al.(2025)Intelligence, Black, Brown, Darpinian,
  Dhabalia, Driess, Esmail, Equi, Finn, Fusai, et~al.]{intelligence2025pi_}
P.~Intelligence, K.~Black, N.~Brown, J.~Darpinian, K.~Dhabalia, D.~Driess,
  A.~Esmail, M.~Equi, C.~Finn, N.~Fusai, et~al.
\newblock $\pi_{0.5}$: a vision-language-action model with open-world
  generalization.
\newblock \emph{arXiv preprint arXiv:2504.16054}, 2025.

\bibitem[Shi et~al.(2025)Shi, Ichter, Equi, Ke, Pertsch, Vuong, Tanner,
  Walling, Wang, Fusai, et~al.]{shi2025hi}
L.~X. Shi, B.~Ichter, M.~Equi, L.~Ke, K.~Pertsch, Q.~Vuong, J.~Tanner,
  A.~Walling, H.~Wang, N.~Fusai, et~al.
\newblock Hi robot: Open-ended instruction following with hierarchical
  vision-language-action models.
\newblock \emph{arXiv preprint arXiv:2502.19417}, 2025.

\bibitem[Li et~al.(2025)Li, Deng, Zhang, Jang, Memmel, Yu, Garrett, Ramos, Fox,
  Li, et~al.]{li2025hamster}
Y.~Li, Y.~Deng, J.~Zhang, J.~Jang, M.~Memmel, R.~Yu, C.~R. Garrett, F.~Ramos,
  D.~Fox, A.~Li, et~al.
\newblock Hamster: Hierarchical action models for open-world robot
  manipulation.
\newblock \emph{arXiv preprint arXiv:2502.05485}, 2025.

\bibitem[Figure(2024)]{figure2024helix}
A.~Figure.
\newblock Helix: A vision-language-action model for generalist humanoid
  control.
\newblock \emph{Figure AI News}, 2024.

\bibitem[Kahneman(2011)]{kahneman2011thinking}
D.~Kahneman.
\newblock Thinking, fast and slow.
\newblock \emph{Farrar, Straus and Giroux}, 2011.

\bibitem[Tan et~al.(2025)Tan, Hao, Chi, Lin, Lyu, Cao, Liang, Chen, Lyu, Peng,
  et~al.]{tan2025roboos}
H.~Tan, X.~Hao, C.~Chi, M.~Lin, Y.~Lyu, M.~Cao, D.~Liang, Z.~Chen, M.~Lyu,
  C.~Peng, et~al.
\newblock Roboos: A hierarchical embodied framework for cross-embodiment and
  multi-agent collaboration.
\newblock \emph{arXiv preprint arXiv:2505.03673}, 2025.

\bibitem[Sutton et~al.(1999)Sutton, Precup, and Singh]{sutton1999between}
R.~S. Sutton, D.~Precup, and S.~Singh.
\newblock Between mdps and semi-mdps: A framework for temporal abstraction in
  reinforcement learning.
\newblock \emph{Artificial intelligence}, 112\penalty0 (1-2):\penalty0
  181--211, 1999.

\bibitem[French(1999)]{french1999catastrophic}
R.~M. French.
\newblock Catastrophic forgetting in connectionist networks.
\newblock \emph{Trends in cognitive sciences}, 3\penalty0 (4):\penalty0
  128--135, 1999.

\bibitem[Brooks(2003)]{brooks2003robust}
R.~Brooks.
\newblock A robust layered control system for a mobile robot.
\newblock \emph{IEEE journal on robotics and automation}, 2\penalty0
  (1):\penalty0 14--23, 2003.

\bibitem[Ahn et~al.(2022)Ahn, Brohan, Brown, Chebotar, Cortes, David, Finn, Fu,
  Gopalakrishnan, Hausman, et~al.]{ahn2022can}
M.~Ahn, A.~Brohan, N.~Brown, Y.~Chebotar, O.~Cortes, B.~David, C.~Finn, C.~Fu,
  K.~Gopalakrishnan, K.~Hausman, et~al.
\newblock Do as i can, not as i say: Grounding language in robotic affordances.
\newblock \emph{arXiv preprint arXiv:2204.01691}, 2022.

\bibitem[Hu et~al.(2025)Hu, Stone, and Mart{\'\i}n-Mart{\'\i}n]{hu2025slac}
J.~Hu, P.~Stone, and R.~Mart{\'\i}n-Mart{\'\i}n.
\newblock Slac: Simulation-pretrained latent action space for whole-body
  real-world rl.
\newblock \emph{arXiv preprint arXiv:2506.04147}, 2025.

\bibitem[Su et~al.(2025)Su, Zhang, Rahmanian, Gao, Liao, Regan, Sreenath, and
  Sastry]{su2025hitter}
Z.~Su, B.~Zhang, N.~Rahmanian, Y.~Gao, Q.~Liao, C.~Regan, K.~Sreenath, and
  S.~S. Sastry.
\newblock Hitter: A humanoid table tennis robot via hierarchical planning and
  learning.
\newblock \emph{arXiv preprint arXiv:2508.21043}, 2025.

\bibitem[Fu et~al.(2026)Fu, Yu, El-Refai, Kou, Xue, Huang, Xiao, Wang, Li, Shi,
  Wu, Sastry, Zhu, Goldberg, and Fan]{Fu2026CaPXAF}
M.~Fu, J.~Yu, K.~El-Refai, E.~Kou, H.~Xue, H.~Huang, W.~Xiao, G.~Wang, F.-F.
  Li, G.~Shi, J.~Wu, S.~Sastry, Y.~Zhu, K.~Goldberg, and L.~J. Fan.
\newblock Cap-x: A framework for benchmarking and improving coding agents for
  robot manipulation.
\newblock 2026.
\newblock URL \url{https://api.semanticscholar.org/CorpusID:286770427}.

\bibitem[Shi et~al.(2025)Shi, Yang, Chao, Wan, Shao, Lei, Qian, Le, Chaudhari,
  Daniilidis, Wen, and Jayaraman]{Shi2025MaestroOR}
J.~Shi, R.~Yang, K.~Chao, S.~Wan, Y.~S. Shao, J.~Lei, J.~Qian, L.~Le,
  P.~Chaudhari, K.~Daniilidis, C.~Wen, and D.~Jayaraman.
\newblock Maestro: Orchestrating robotics modules with vision-language models
  for zero-shot generalist robots.
\newblock \emph{ArXiv}, abs/2511.00917, 2025.
\newblock URL \url{https://api.semanticscholar.org/CorpusID:282738665}.

\bibitem[Jiang et~al.(2025)Jiang, Yuan, Liu, Lu, Cui, Liu, Cheng, Gao, Xu, and
  Zhao]{jiang2025galaxeaopenworlddatasetg0}
T.~Jiang, T.~Yuan, Y.~Liu, C.~Lu, J.~Cui, X.~Liu, S.~Cheng, J.~Gao, H.~Xu, and
  H.~Zhao.
\newblock Galaxea open-world dataset and g0 dual-system vla model, 2025.
\newblock URL \url{https://arxiv.org/abs/2509.00576}.

\bibitem[Ding et~al.(2025)Ding, Ma, Tong, Zou, Luo, Fan, Wang, Lu, Mo, Liu,
  et~al.]{ding2025humanoid}
P.~Ding, J.~Ma, X.~Tong, B.~Zou, X.~Luo, Y.~Fan, T.~Wang, H.~Lu, P.~Mo, J.~Liu,
  et~al.
\newblock Humanoid-vla: Towards universal humanoid control with visual
  integration.
\newblock \emph{arXiv preprint arXiv:2502.14795}, 2025.

\bibitem[Comanici et~al.(2025)Comanici, Bieber, Schaekermann, Pasupat,
  Sachdeva, Dhillon, Blistein, Ram, Zhang, Rosen, et~al.]{comanici2025gemini}
G.~Comanici, E.~Bieber, M.~Schaekermann, I.~Pasupat, N.~Sachdeva, I.~Dhillon,
  M.~Blistein, O.~Ram, D.~Zhang, E.~Rosen, et~al.
\newblock Gemini 2.5: Pushing the frontier with advanced reasoning,
  multimodality, long context, and next generation agentic capabilities.
\newblock \emph{arXiv preprint arXiv:2507.06261}, 2025.

\bibitem[Intelligence et~al.(2026)Intelligence, Ai, Amin, Aniceto, Balakrishna,
  Balke, Black, Bokinsky, Cao, Charbonnier, Choudhary, Collins, Conley,
  Connors, Darpinian, Dhabalia, Dhaka, DiCarlo, Driess, Equi, Esmail, Fang,
  Finn, Glossop, Godden, Goryachev, Groom, Habeeb, Hancock, Hausman, Hussein,
  Hwang, Ichter, Jacobsen, Jakubczak, Jen, Jones, Kammerer, Katz, Ke, Khadikov,
  Kuchi, Lamb, LeBlanc, LeCount, Levine, Li, Li-Bell, Lialin, Liang, Lim, Lu,
  Luo, Mano, Marwaha, Mongush, Murphy, Nair, Patterson, Pertsch, Ren, Schelske,
  Sharma, Shi, Shi, Smith, Springenberg, Stachowicz, Stoeckle, Tang, Tanner,
  Tekeste, Torne, Vedder, Vuong, Walling, Wang, Wang, Wang, Whalen, Whitmore,
  Williams, Xu, Yoo, Yu, Zhang, Zhang, and
  Zhilinsky]{intelligence2026pi07steerablegeneralistrobotic}
P.~Intelligence, B.~Ai, A.~Amin, R.~Aniceto, A.~Balakrishna, G.~Balke,
  K.~Black, G.~Bokinsky, S.~Cao, T.~Charbonnier, V.~Choudhary, F.~Collins,
  K.~Conley, G.~Connors, J.~Darpinian, K.~Dhabalia, M.~Dhaka, J.~DiCarlo,
  D.~Driess, M.~Equi, A.~Esmail, Y.~Fang, C.~Finn, C.~Glossop, T.~Godden,
  I.~Goryachev, L.~Groom, H.~Habeeb, H.~Hancock, K.~Hausman, G.~Hussein,
  V.~Hwang, B.~Ichter, C.~Jacobsen, S.~Jakubczak, R.~Jen, T.~Jones,
  G.~Kammerer, B.~Katz, L.~Ke, M.~Khadikov, C.~Kuchi, M.~Lamb, D.~LeBlanc,
  B.~LeCount, S.~Levine, X.~Li, A.~Li-Bell, V.~Lialin, Z.~Liang, W.~Lim, Y.~Lu,
  E.~Luo, V.~Mano, N.~Marwaha, A.~Mongush, L.~Murphy, S.~Nair, T.~Patterson,
  K.~Pertsch, A.~Z. Ren, G.~Schelske, C.~Sharma, B.~Shi, L.~X. Shi, L.~Smith,
  J.~T. Springenberg, K.~Stachowicz, W.~Stoeckle, J.~Tang, J.~Tanner,
  S.~Tekeste, M.~Torne, K.~Vedder, Q.~Vuong, A.~Walling, H.~Wang, J.~Wang,
  X.~Wang, C.~Whalen, S.~Whitmore, B.~Williams, C.~Xu, S.~Yoo, L.~Yu, W.~Zhang,
  Z.~Zhang, and U.~Zhilinsky.
\newblock ${\pi}_{0.7}$: a steerable generalist robotic foundation model with
  emergent capabilities, 2026.
\newblock URL \url{https://arxiv.org/abs/2604.15483}.

\bibitem[Huang et~al.(2026)Huang, Shao, Wang, Chen, Sun, Guo, Schwager, and
  Bohg]{Huang2026BreakingLP}
S.~Huang, J.~Shao, K.~Wang, Q.~Chen, J.~Sun, Y.~Guo, M.~Schwager, and J.~Bohg.
\newblock Breaking lock-in: Preserving steerability under low-data vla
  post-training.
\newblock 2026.
\newblock URL \url{https://api.semanticscholar.org/CorpusID:287777164}.

\bibitem[Chen et~al.(2026)Chen, Bhatia, Glossop, Mathihalli, Doshi, Tang,
  Driess, Pertsch, and Levine]{Chen2026SteerableVP}
W.~Chen, J.~S. Bhatia, C.~Glossop, N.~Mathihalli, R.~Doshi, A.~Tang, D.~Driess,
  K.~Pertsch, and S.~Levine.
\newblock Steerable vision-language-action policies for embodied reasoning and
  hierarchical control.
\newblock \emph{ArXiv}, abs/2602.13193, 2026.
\newblock URL \url{https://api.semanticscholar.org/CorpusID:285606737}.

\bibitem[Du et~al.(2023)Du, Konyushkova, Denil, Raju, Landon, Hill, De~Freitas,
  and Cabi]{du2023vision}
Y.~Du, K.~Konyushkova, M.~Denil, A.~Raju, J.~Landon, F.~Hill, N.~De~Freitas,
  and S.~Cabi.
\newblock Vision-language models as success detectors.
\newblock \emph{arXiv preprint arXiv:2303.07280}, 2023.

\bibitem[Majumdar et~al.(2024)Majumdar, Ajay, Zhang, Putta, Yenamandra, Henaff,
  Silwal, Mcvay, Maksymets, Arnaud, et~al.]{majumdar2024openeqa}
A.~Majumdar, A.~Ajay, X.~Zhang, P.~Putta, S.~Yenamandra, M.~Henaff, S.~Silwal,
  P.~Mcvay, O.~Maksymets, S.~Arnaud, et~al.
\newblock Openeqa: Embodied question answering in the era of foundation models.
\newblock In \emph{Proceedings of the IEEE/CVF conference on computer vision
  and pattern recognition}, pages 16488--16498, 2024.

\bibitem[Park et~al.(2023)Park, O'Brien, Cai, Morris, Liang, and
  Bernstein]{park2023generative}
J.~S. Park, J.~O'Brien, C.~J. Cai, M.~R. Morris, P.~Liang, and M.~S. Bernstein.
\newblock Generative agents: Interactive simulacra of human behavior.
\newblock In \emph{Proceedings of the 36th annual acm symposium on user
  interface software and technology}, pages 1--22, 2023.

\bibitem[Shinn et~al.(2023)Shinn, Cassano, Gopinath, Narasimhan, and
  Yao]{shinn2023reflexion}
N.~Shinn, F.~Cassano, A.~Gopinath, K.~Narasimhan, and S.~Yao.
\newblock Reflexion: Language agents with verbal reinforcement learning.
\newblock \emph{Advances in Neural Information Processing Systems},
  36:\penalty0 8634--8652, 2023.

\bibitem[Bai et~al.(2025)]{bai2025qwen3vl}
S.~Bai et~al.
\newblock Qwen3-vl technical report.
\newblock \emph{arXiv preprint arXiv:2511.21631}, 2025.

\bibitem[Liu et~al.(2023)Liu, Zhu, Gao, Feng, Liu, Zhu, and
  Stone]{liu2023libero}
B.~Liu, Y.~Zhu, C.~Gao, Y.~Feng, Q.~Liu, Y.~Zhu, and P.~Stone.
\newblock {LIBERO}: Benchmarking knowledge transfer for lifelong robot
  learning.
\newblock \emph{Advances in Neural Information Processing Systems}, 36, 2023.

\bibitem[Hu et~al.(2026)Hu, Shim, Tang, Sung, Liu, Stone, and
  Martin-Martin]{hu2026simple}
J.~Hu, J.~Shim, C.~Tang, Y.~Sung, B.~Liu, P.~Stone, and R.~Martin-Martin.
\newblock Simple recipe works: Vision-language-action models are natural
  continual learners with reinforcement learning, 2026.
\newblock URL \url{https://arxiv.org/abs/2603.11653}.

\bibitem[Radford et~al.(2021)Radford, Kim, Hallacy, Ramesh, Goh, Agarwal,
  Sastry, Askell, Mishkin, Clark, et~al.]{radford2021learning}
A.~Radford, J.~W. Kim, C.~Hallacy, A.~Ramesh, G.~Goh, S.~Agarwal, G.~Sastry,
  A.~Askell, P.~Mishkin, J.~Clark, et~al.
\newblock Learning transferable visual models from natural language
  supervision.
\newblock In \emph{International conference on machine learning}, pages
  8748--8763. PmLR, 2021.

\bibitem[Alayrac et~al.(2022)Alayrac, Donahue, Luc, Miech, Barr, Hasson, Lenc,
  Mensch, Millican, Reynolds, et~al.]{alayrac2022flamingo}
J.-B. Alayrac, J.~Donahue, P.~Luc, A.~Miech, I.~Barr, Y.~Hasson, K.~Lenc,
  A.~Mensch, K.~Millican, M.~Reynolds, et~al.
\newblock Flamingo: a visual language model for few-shot learning.
\newblock \emph{Advances in neural information processing systems},
  35:\penalty0 23716--23736, 2022.

\bibitem[Li et~al.(2023)Li, Li, Savarese, and Hoi]{li2023blip}
J.~Li, D.~Li, S.~Savarese, and S.~Hoi.
\newblock Blip-2: Bootstrapping language-image pre-training with frozen image
  encoders and large language models.
\newblock In \emph{International conference on machine learning}, pages
  19730--19742. PMLR, 2023.

\bibitem[Team et~al.(2023)Team, Anil, Borgeaud, Alayrac, Yu, Soricut,
  Schalkwyk, Dai, Hauth, Millican, et~al.]{team2023gemini}
G.~Team, R.~Anil, S.~Borgeaud, J.-B. Alayrac, J.~Yu, R.~Soricut, J.~Schalkwyk,
  A.~M. Dai, A.~Hauth, K.~Millican, et~al.
\newblock Gemini: a family of highly capable multimodal models.
\newblock \emph{arXiv preprint arXiv:2312.11805}, 2023.

\bibitem[Beyer et~al.(2024)Beyer, Steiner, Pinto, Kolesnikov, Wang, Salz,
  Neumann, Alabdulmohsin, Tschannen, Bugliarello, et~al.]{beyer2024paligemma}
L.~Beyer, A.~Steiner, A.~S. Pinto, A.~Kolesnikov, X.~Wang, D.~Salz, M.~Neumann,
  I.~Alabdulmohsin, M.~Tschannen, E.~Bugliarello, et~al.
\newblock Paligemma: A versatile 3b vlm for transfer.
\newblock \emph{arXiv preprint arXiv:2407.07726}, 2024.

\bibitem[Bai et~al.(2023)Bai, Bai, Chu, Cui, Dang, Deng, Fan, Ge, Han, Huang,
  et~al.]{bai2023qwen}
J.~Bai, S.~Bai, Y.~Chu, Z.~Cui, K.~Dang, X.~Deng, Y.~Fan, W.~Ge, Y.~Han,
  F.~Huang, et~al.
\newblock Qwen technical report.
\newblock \emph{arXiv preprint arXiv:2309.16609}, 2023.

\bibitem[Zhu et~al.(2023)Zhu, Chen, Shen, Li, and Elhoseiny]{zhu2023minigpt}
D.~Zhu, J.~Chen, X.~Shen, X.~Li, and M.~Elhoseiny.
\newblock Minigpt-4: Enhancing vision-language understanding with advanced
  large language models.
\newblock \emph{arXiv preprint arXiv:2304.10592}, 2023.

\bibitem[Driess et~al.(2023)Driess, Xia, Sajjadi, Lynch, Chowdhery, Wahid,
  Tompson, Vuong, Yu, Huang, et~al.]{driess2023palm}
D.~Driess, F.~Xia, M.~S. Sajjadi, C.~Lynch, A.~Chowdhery, A.~Wahid, J.~Tompson,
  Q.~Vuong, T.~Yu, W.~Huang, et~al.
\newblock Palm-e: An embodied multimodal language model.
\newblock \emph{arXiv preprint}, 2023.

\bibitem[Liu et~al.(2023)Liu, Li, Wu, and Lee]{liu2023visual}
H.~Liu, C.~Li, Q.~Wu, and Y.~J. Lee.
\newblock Visual instruction tuning.
\newblock \emph{Advances in neural information processing systems},
  36:\penalty0 34892--34916, 2023.

\bibitem[Zhong et~al.(2025)Zhong, Bai, Cai, Huang, Chen, Zhang, Wang, Guo,
  Guan, Lui, Qi, Liang, Chen, and
  Yang]{zhong2025surveyvisionlanguageactionmodelsaction}
Y.~Zhong, F.~Bai, S.~Cai, X.~Huang, Z.~Chen, X.~Zhang, Y.~Wang, S.~Guo,
  T.~Guan, K.~N. Lui, Z.~Qi, Y.~Liang, Y.~Chen, and Y.~Yang.
\newblock A survey on vision-language-action models: An action tokenization
  perspective, 2025.
\newblock URL \url{https://arxiv.org/abs/2507.01925}.

\bibitem[Sapkota et~al.(2025)Sapkota, Cao, Roumeliotis, and
  Karkee]{sapkota2025visionlanguageactionmodelsconceptsprogress}
R.~Sapkota, Y.~Cao, K.~I. Roumeliotis, and M.~Karkee.
\newblock Vision-language-action models: Concepts, progress, applications and
  challenges, 2025.
\newblock URL \url{https://arxiv.org/abs/2505.04769}.

\bibitem[Ma et~al.(2025)Ma, Song, Zhuang, Hao, and
  King]{ma2025surveyvisionlanguageactionmodelsembodied}
Y.~Ma, Z.~Song, Y.~Zhuang, J.~Hao, and I.~King.
\newblock A survey on vision-language-action models for embodied ai, 2025.
\newblock URL \url{https://arxiv.org/abs/2405.14093}.

\bibitem[Guruprasad et~al.(2024)Guruprasad, Sikka, Song, Wang, and
  Liang]{guruprasad2024benchmarkingvisionlanguage}
P.~Guruprasad, H.~Sikka, J.~Song, Y.~Wang, and P.~P. Liang.
\newblock Benchmarking vision, language, and action models on robotic learning
  tasks, 2024.
\newblock URL \url{https://arxiv.org/abs/2411.05821}.

\bibitem[Gao et~al.(2025)Gao, Belkhale, Dasari, Balakrishna, Shah, and
  Sadigh]{gao2025taxonomy}
J.~Gao, S.~Belkhale, S.~Dasari, A.~Balakrishna, D.~Shah, and D.~Sadigh.
\newblock A taxonomy for evaluating generalist robot policies.
\newblock \emph{arXiv preprint arXiv:2503.01238}, 2025.

\bibitem[Li et~al.(2024)Li, Li, Liu, Wang, Liu, Kang, Ma, Kong, Zhang, and
  Liu]{li2024generalistrobotpoliciesmatters}
X.~Li, P.~Li, M.~Liu, D.~Wang, J.~Liu, B.~Kang, X.~Ma, T.~Kong, H.~Zhang, and
  H.~Liu.
\newblock Towards generalist robot policies: What matters in building
  vision-language-action models, 2024.
\newblock URL \url{https://arxiv.org/abs/2412.14058}.

\bibitem[Gao et~al.(2025)Gao, Liu, Chi, Huang, Fei, Hou, Zhang, Lin, Fang,
  Jiang, and Shao]{gao2025vlaosstructuringdissectingplanning}
C.~Gao, Z.~Liu, Z.~Chi, J.~Huang, X.~Fei, Y.~Hou, Y.~Zhang, Y.~Lin, Z.~Fang,
  Z.~Jiang, and L.~Shao.
\newblock Vla-os: Structuring and dissecting planning representations and
  paradigms in vision-language-action models, 2025.
\newblock URL \url{https://arxiv.org/abs/2506.17561}.

\end{thebibliography}

\clearpage
\setcounter{page}{1}
% \maketitlesupplementary
% \appendix

% \section*{Appendix}
% \addcontentsline{toc}{section}{Appendix} % optional
\beginSupplementaryMaterials

\appendix

\begin{figure*}[h]
\centering

\begin{subfigure}{0.24\textwidth}
    \includegraphics[width=\linewidth]{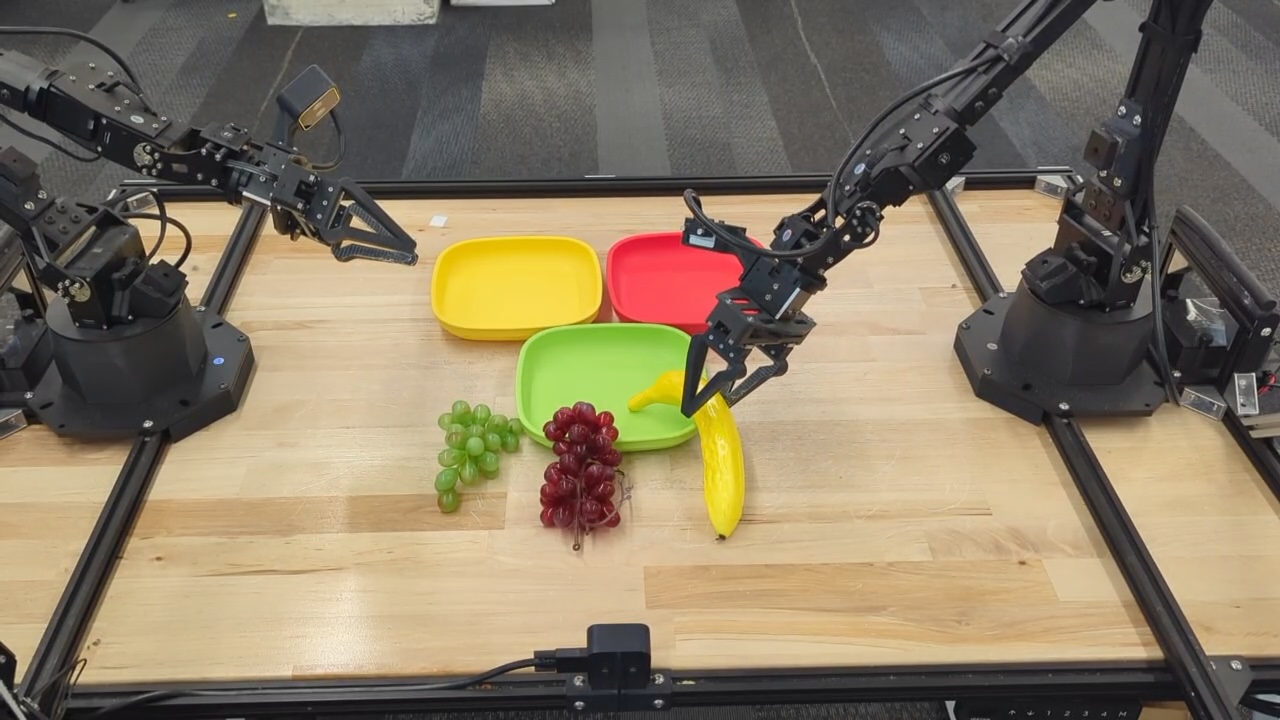}
    \caption{Step 1}
\end{subfigure}
\hfill
\begin{subfigure}{0.24\textwidth}
    \includegraphics[width=\linewidth]{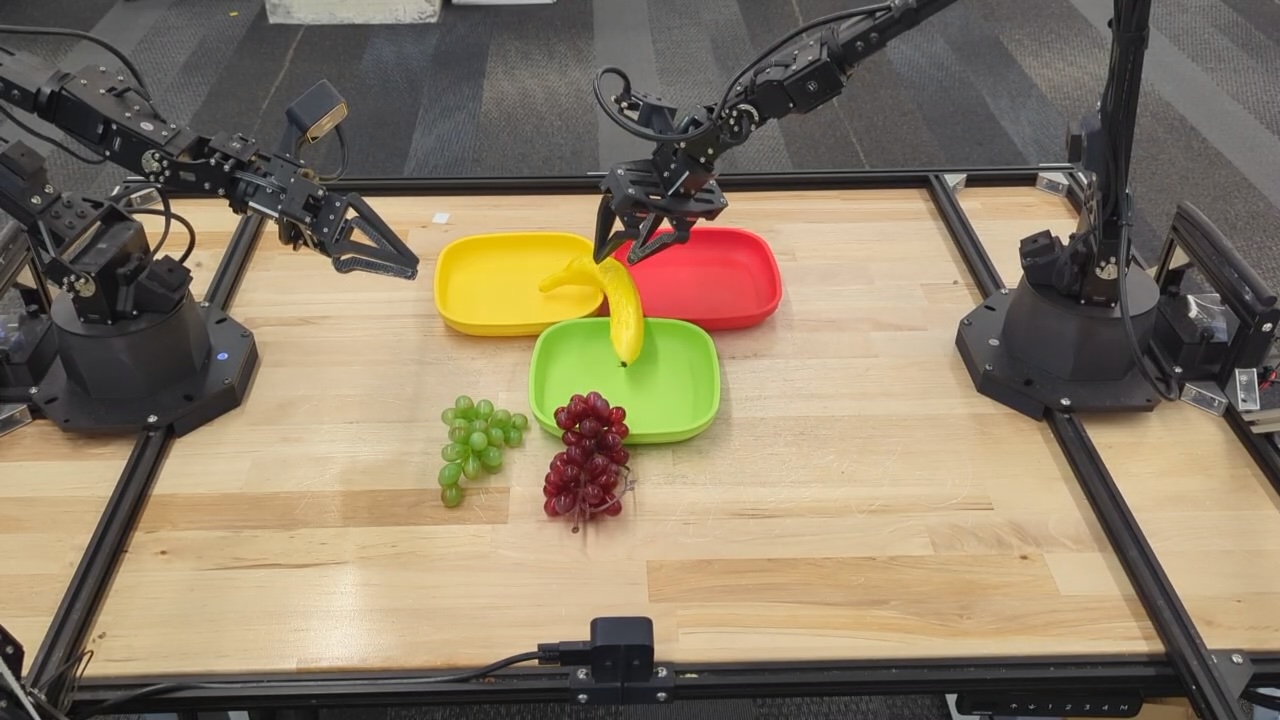}
    \caption{Step 2}
\end{subfigure}
\hfill
\begin{subfigure}{0.24\textwidth}
    \includegraphics[width=\linewidth]{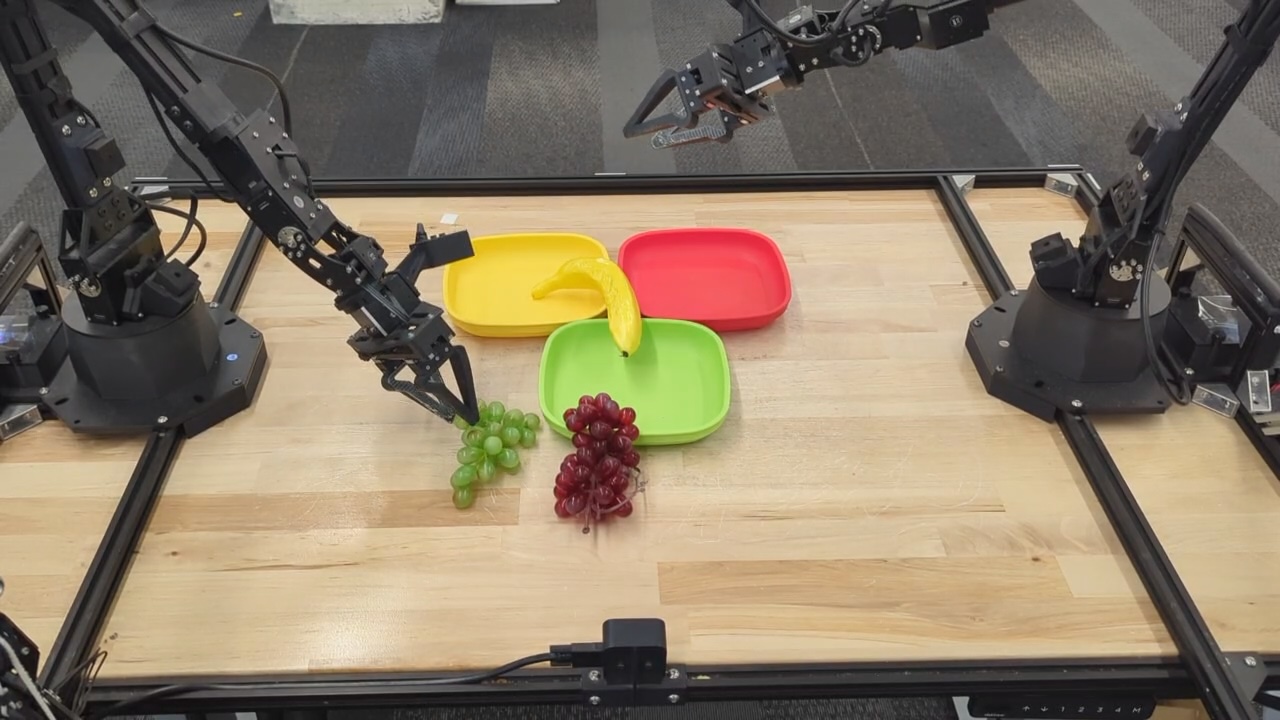}
    \caption{Step 3}
\end{subfigure}
\hfill
\begin{subfigure}{0.24\textwidth}
    \includegraphics[width=\linewidth]{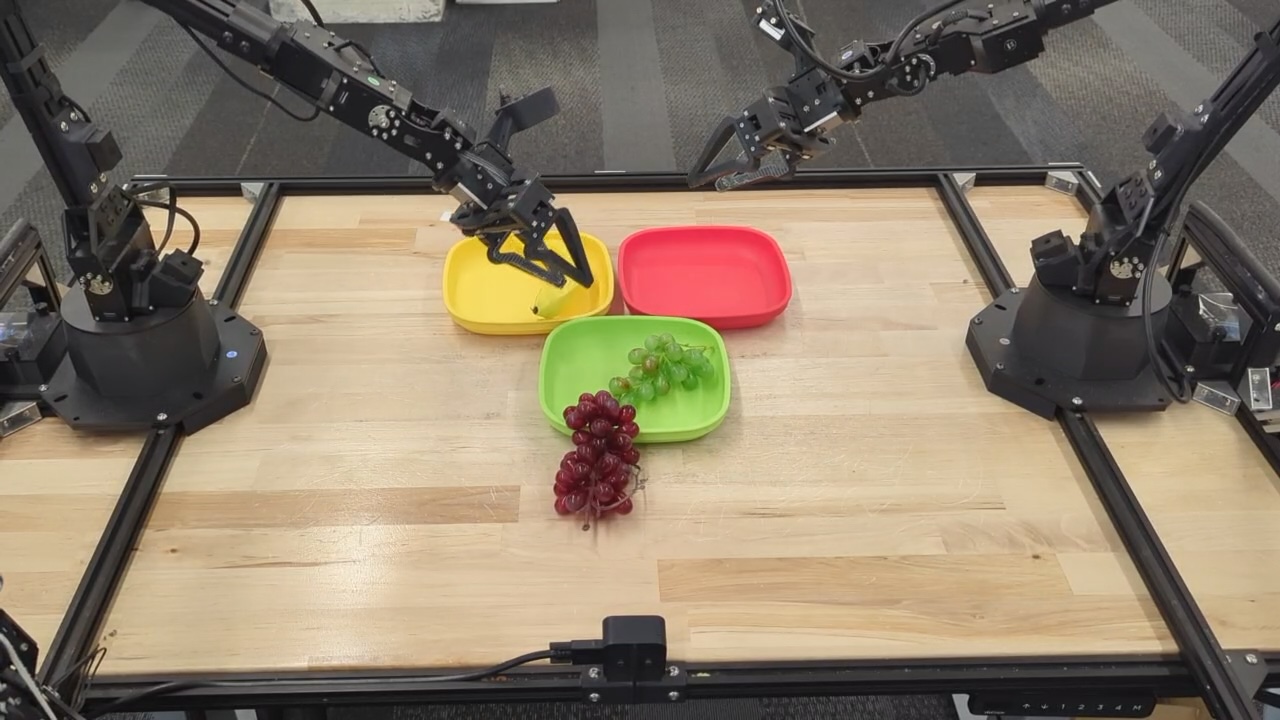}
    \caption{Step 4}
\end{subfigure}
\hfill
\begin{subfigure}{0.24\textwidth}
    \includegraphics[width=\linewidth]{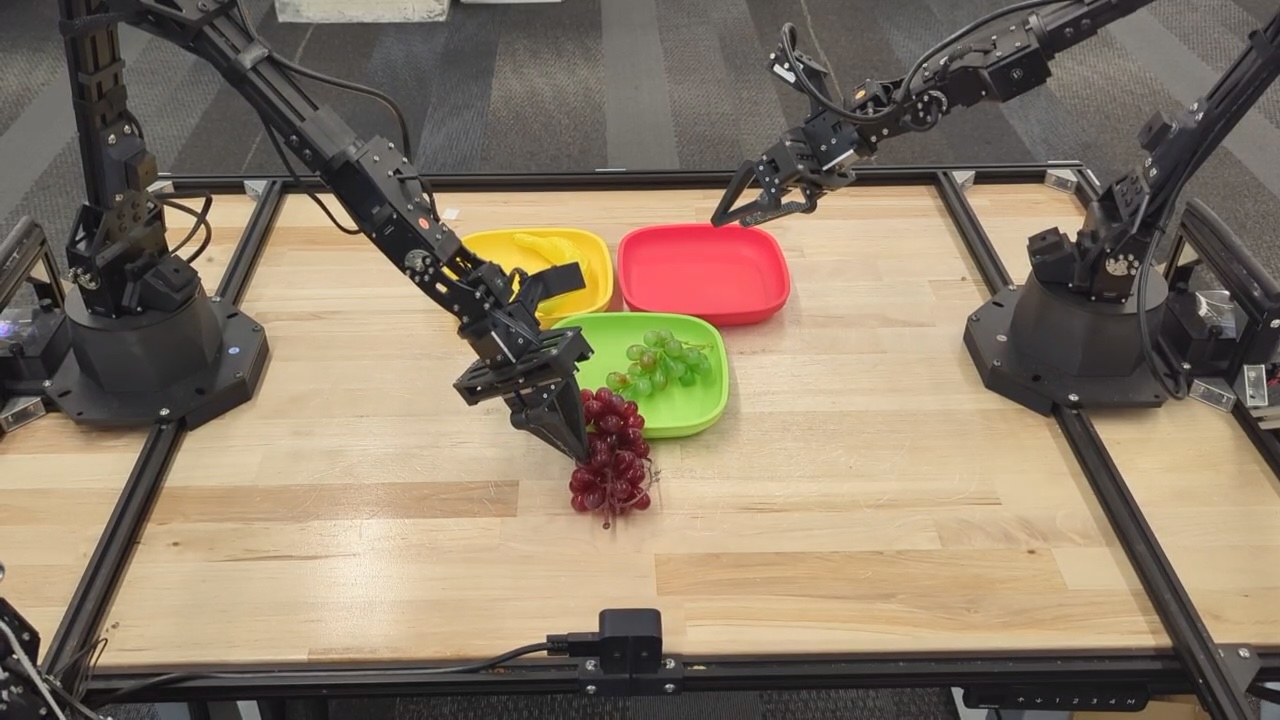}
    \caption{Step 5}
\end{subfigure}
\hfill
\begin{subfigure}{0.24\textwidth}
    \includegraphics[width=\linewidth]{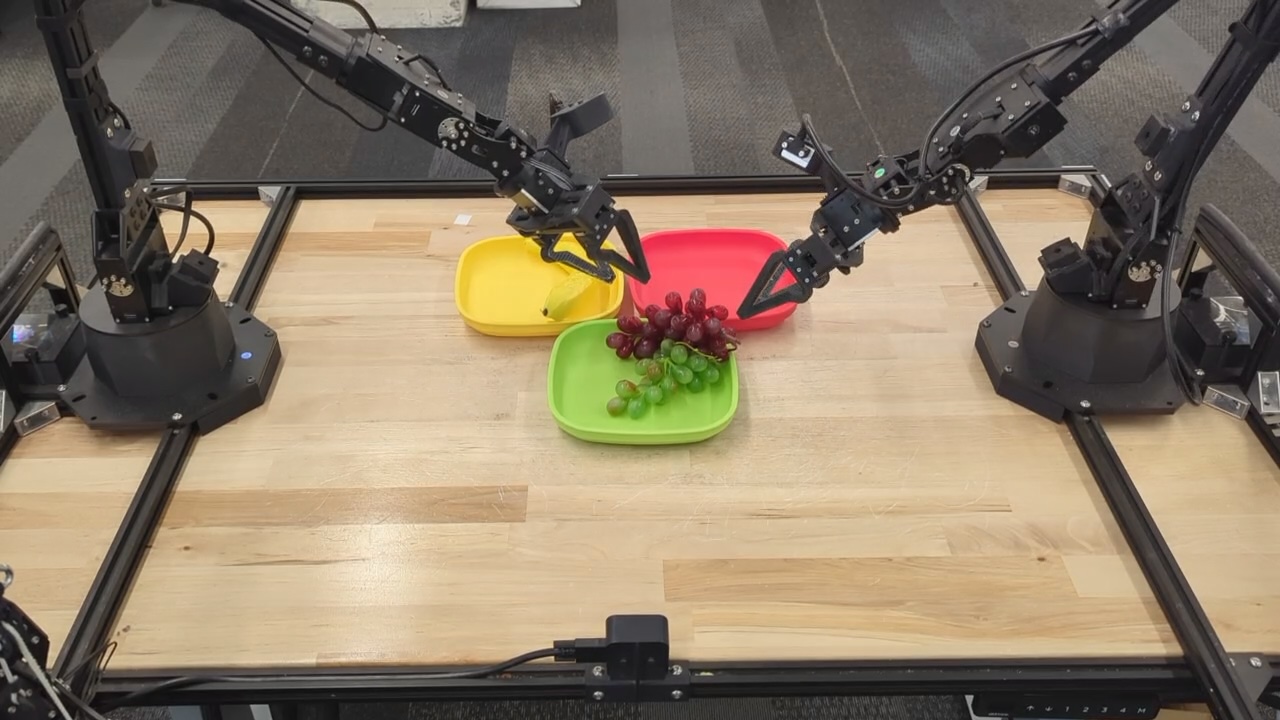}
    \caption{Step 6}
\end{subfigure}
\hfill
\begin{subfigure}{0.24\textwidth}
    \includegraphics[width=\linewidth]{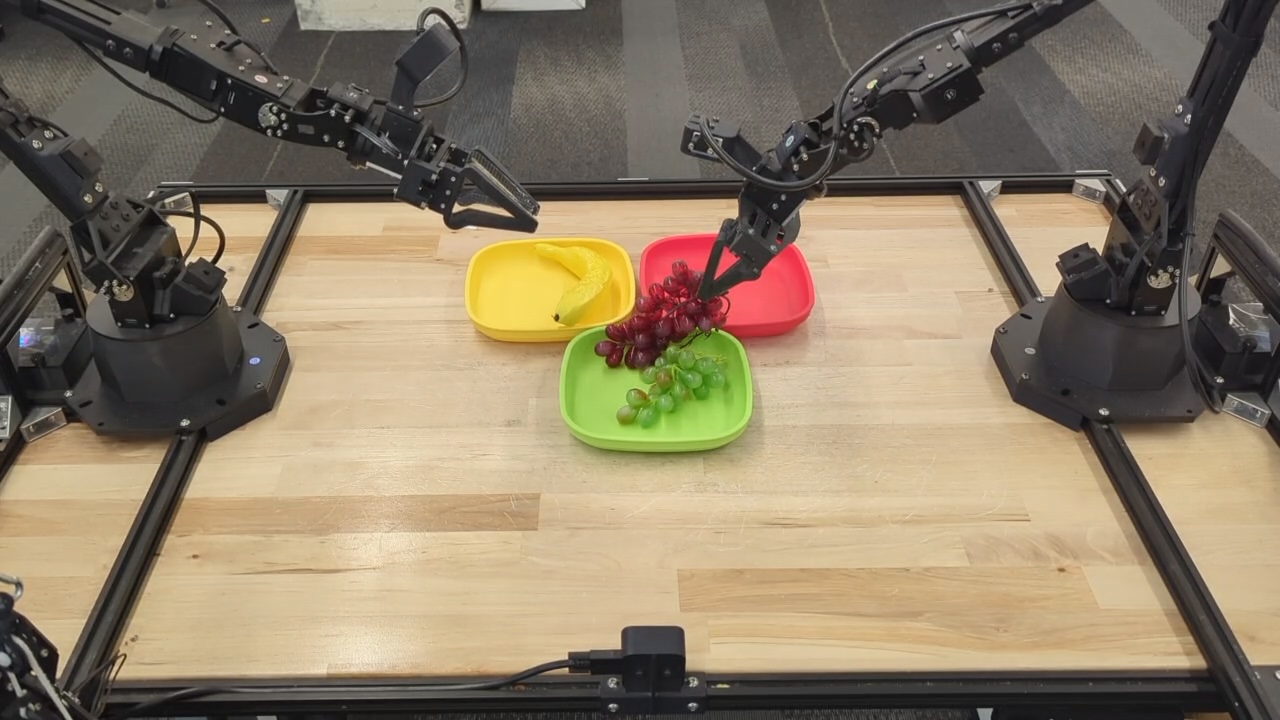}
    \caption{Step 7}
\end{subfigure}
\hfill
\begin{subfigure}{0.24\textwidth}
    \includegraphics[width=\linewidth]{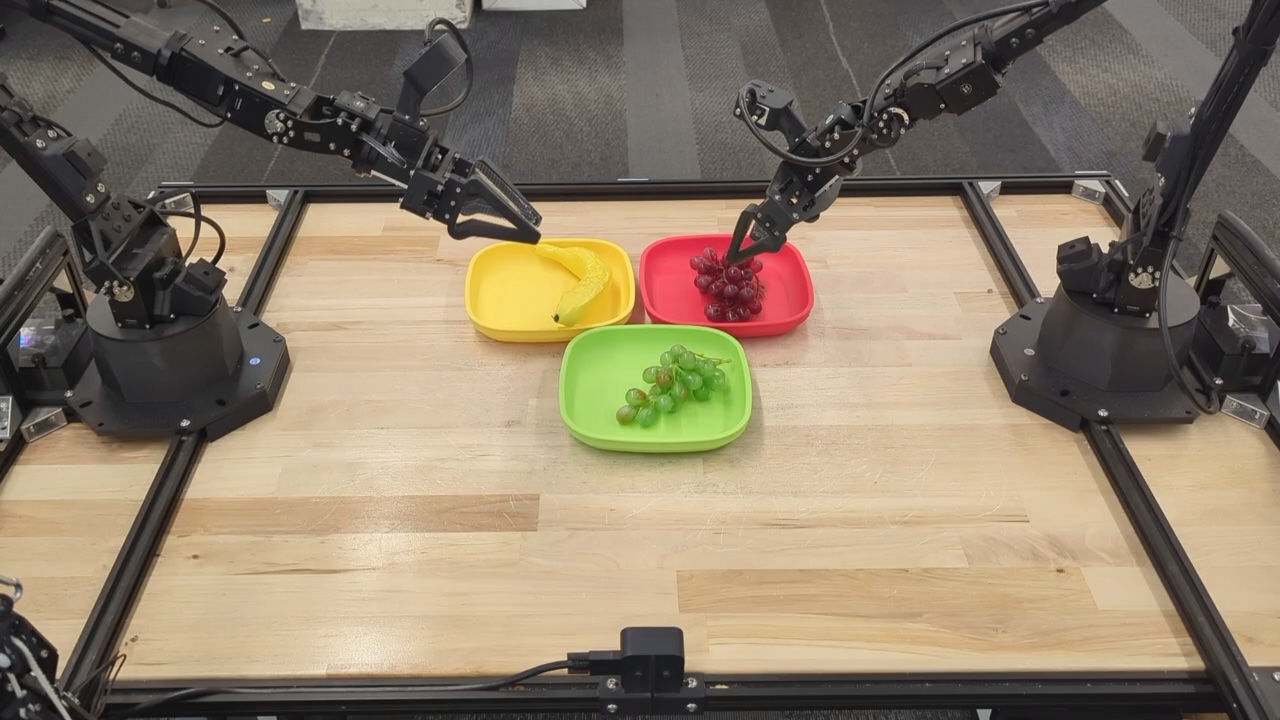}
    \caption{Step 8}
\end{subfigure}

\caption{Motion sequence of the real ALOHA robot. Orchestration allows the robot to recover from the fruit misplacement (step 6) and eventually solve the task (step 8).}
\label{fig:motion_sequence}
\end{figure*}

\section{Evaluation Setup}
\label{sup:eval}

We carry out our main experiments in the MuJoCo Aloha suite, and show some of these environments in Fig.~\ref{fig:four_subfigures}.
Operating in simulation brings us two main advantages: first, simulation parallelization allows us to run large-scale evaluations and obtain results with statistical significance. Second, we can leverage privileged information available in simulation to examine counterfactual hypotheses that can suggest directions for future improvement.

Ideally, a good hierarchical VLA system should have the following desired abilities:
\begin{itemize}
    \item The high-level VLM has the ability to learn about the language affordance of the VLA from experiences, and eventually, only generate commands that respect the affordance.
    \item The system can tackle long-horizon tasks by breaking it down into executable sub-tasks.
    \item The system is able to interpret and reason about indirect instructions by leveraging the strong prior from the VLM.
\end{itemize}

In an effort to disentangle these different capabilities, we categorize our evaluation tasks into three different categories, such that each category of tasks roughly corresponds to one desired capability, allowing us to examine these capabilities separately.
We describe the categorization below and discuss the tasks in detail in Appendix~\ref{sup:task}.

\begin{itemize}
    \item \textbf{Short-horizon}: the task has length similar to the VLA training trajectories, such as pick and place of a single object.
    \item \textbf{Long-horizon}: the task length is significantly longer than the VLA training trajectories, requiring non-trivial compositions of short-horizon skills.
    \item \textbf{Reasoning}: instructions need interpretation, e.g. ``put the object you pour coffee in on plate''.
\end{itemize}

In our experiments, the result for each task category (i.e. short-horizon, long-horizon, reasoning) is averaged over 5 tasks and 200 independent trials per task. All tasks are given a two-minute budget per trial. The error bar represents the standard error arising from finite-sample binomial uncertainty. 
Whenever we are examining one component, we fix the rest of the components to a \emph{standard} setup for result consistency, as described below:
\begin{itemize}
    \item The VLM is set to be Gemini 2.5 Flash with thinking on.
    \item The VLA is set to be the GROD 3B model trained only on real robot data.
    \item The termination condition is set to be fixed frequency switch with a duration of 8 seconds.
    \item The observation representation is set to be scene description with contact information.
    \item The memory window is set to be 3, without using any summarization.
\end{itemize}

Unless otherwise noted, we use Gemini 2.5 Flash with thinking enabled for queries such as scene descriptions and termination generation. 

% \chadded[id=rebuttal]{\textbf{Orchestration cost.} We quantify the orchestration cost by averaging over evaluation episodes. The Naive Hierarchy makes 6.2 VLM calls per episode, whereas the Best Hierarchy makes 23.2 total calls and generates approximately 9,483 tokens. At our measured Gemini 2.5 Flash throughput of 212.5 tokens/s, this corresponds to approximately 44.6 seconds of VLM generation per episode. This overhead is non-negligible, although it remains below the two-minute robot-execution budget, and should be considered when trading computation against task success.}

\section{Effect of Improving VLA Action Quality}
\label{sup:future}

\begin{wrapfigure}{r}{0.45\textwidth}
    \vspace{-0.3cm}
    \includegraphics[width=0.45\textwidth]{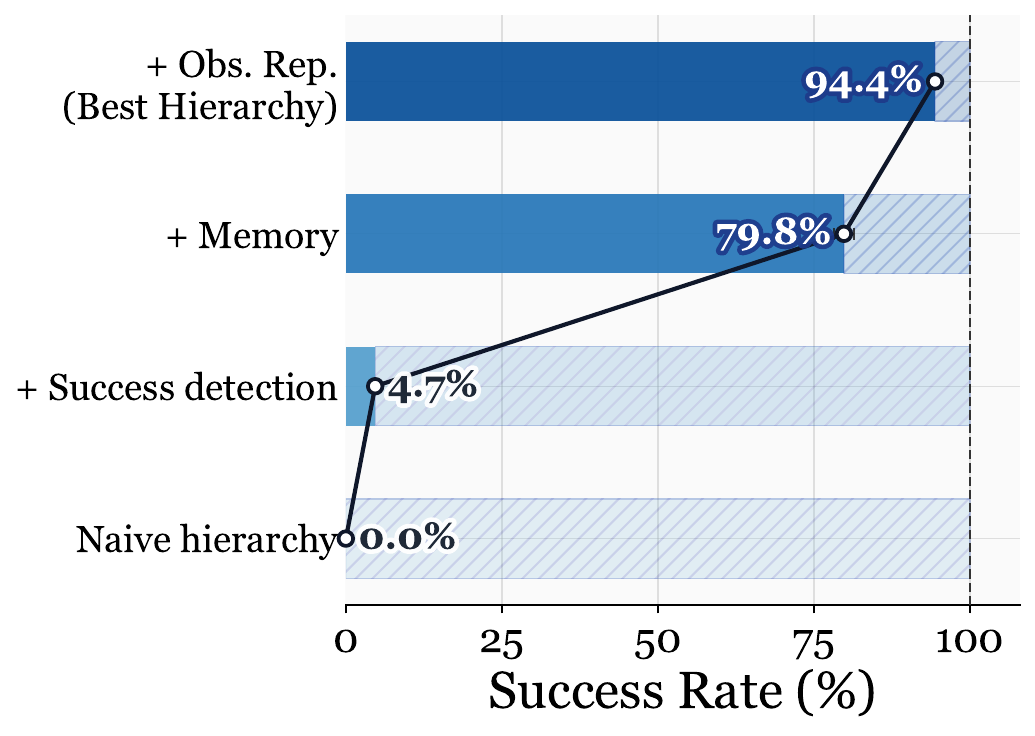}
    \caption{\small Performance of different hierarchies with a scripted low-level policy.}
    \label{fig:scripted}
    \vspace{-0.3cm}
\end{wrapfigure}
In this section, we experiment on how potential improvements in the VLA's action predictions may affect our conclusions.

First of all, note that for a ``perfect'' VLA, hierarchical systems are almost meaningless, since it should be able to directly complete any given task without orchestration.
However, we believe that a more realistic future VLA would be one that can complete a range of \emph{short-horizon} commands with high accuracy, but not necessarily longer or reasoning-based ones.

As a proxy for such VLAs, we created a low-level controller in the form of a scripted policy that utilizes privileged information in the simulator to take actions, such that it can nearly perfectly complete tasks when conditioned on the \textit{right} language command, but would do nothing when it cannot parse the command. More specifically, we parse each generated sentence with spaCy and execute the command only when pre-selected verbs and nouns are successfully extracted. This parser applies only to the scripted future-VLA proxy; the learned VLAs in our main experiments receive generated language directly and do not use a discrete command parser. We test four different settings: the full hierarchical system, removing the observation description, removing the memory, and finally the naive hierarchical system, and show the average performances on a set of challenging long-horizon tasks in Fig.~\ref{fig:scripted}.

% Given that the capabilities of the low-level VLA makes a big difference to the eventual performance, we further explore the setting where we have a VLA that can perfectly generate low-level actions when conditioned on the right language commands.

We can see that the full hierarchical system achieves a very high average success rate of around 95\%.
% , indicating that the VLA low-level acting capability is indeed one of the biggest bottlenecks for existing systems. However, it is also worth noticing that 
however, ablations of hierarchical components (e.g., observation representation, memory, or naive orchestration) can degrade performance from ~95\% success to nearly 0\%. This result suggests that as VLA capabilities improve, hierarchical design and orchestration will remain an important factor, rather than being obviated by better low-level policies.

\section{Robustness to Imperfect Success Detectors}
\label{sup:succ_rob}

In this section, we detail the experiments we conducted for testing the robustness of success detection as a termination criteria under accuracy deterioration. 

We consider this problem from two settings: increasing false positive rate and increasing false negative rate. For each setting, we consider a corruption probability of 10\%, 30\%, and 50\% respectively, and report the results in Fig.~\ref{fig:succ_fp}.
% \jiaheng{So the expected increase is k * j / (1-j). Meaning it is at most 50 steps. That is just negligible...}

Intriguingly, a small amount of success detection error (10\%) does not hurt performance at all, and in fact slightly boost it, suggesting that success detection can be a robust termination condition for hierarchical VLAs. As the detection error rate goes up, however, false positive errors start to drastically impact the system performance, likely because of the false ``command completed'' feedback that causes the VLM to move on to a new command.

While the impact of false negatives (FN) error remains moderate, it is important to note that this result may stem from the fact that our corruption is independent across different states. In other words, even if the current success check failed due to FN, subsequent success checks at later timesteps will still have a good chance of correctly terminating the command. 
By contrast, a success detector in the real world may show high correlation across detection error of consecutive states, meaning that once a FN occurs, the command may fail to terminate for a very long time. As we have shown in the ``execution horizon'' experiment from the previous paragraphs, such a behavior can actually hurt the performance quite significantly.

% As a result, our experimental result here for false-negative perturbation may not fully match what would happen in the real world. 

\section{Discussion of Previous Work on Flat VLAs}
\label{rw:flat}

% \subsection{Vision-Language-Action Models}
\textbf{Vision-Language-Action Models} Internet-scale multi-modal data and transformer-based architectures have given rise to models capable of jointly processing visual and linguistic information, commonly known as vision-language models~\cite{radford2021learning, alayrac2022flamingo, li2023blip, team2023gemini, beyer2024paligemma, bai2023qwen, zhu2023minigpt, driess2023palm, liu2023visual}. 
Building upon these foundation models, the robotics community recently introduced the notion of vision-language-action models, including RT-2 \cite{zitkovich2023rt}, Pi \cite{black2024pi_0}, OpenVLA \cite{kim2024openvla}, Gemini Robotics~\cite{team2025gemini}, and more. These models are fine-tuned from VLMs to map natural-language instructions and perceptual inputs directly to robot actions, allowing VLMs to ``speak the language of robotics'' and ground their extensive knowledge in the physical, embodied world. This adaptation unlocks a paradigm shift in robot learning, enabling unprecedented levels of zero-shot generalization across varied tasks and environments. These VLA models serve as the foundation for the study of this work.

% \subsection{Benchmarks and Evaluations of VLA Systems}
\textbf{Benchmarks and Evaluations of VLA Systems} VLAs have quickly gained traction ever since their introduction, due to the promise of generalization across tasks and embodiments. This increased attention led to surveys \cite{zhong2025surveyvisionlanguageactionmodelsaction, sapkota2025visionlanguageactionmodelsconceptsprogress, ma2025surveyvisionlanguageactionmodelsembodied}, benchmarks \cite{guruprasad2024benchmarkingvisionlanguage, gao2025taxonomy}, and studies \cite{li2024generalistrobotpoliciesmatters, gao2025vlaosstructuringdissectingplanning} that seeks to assess and systematize these advances. However, existing works primarily focus on \textit{flat} VLAs, where a single, monolithic model directly maps instructions to low-level robot actions. By contrast, our work focuses on systematic dissection and evaluation of \textit{hierarchical} VLA systems, where a high-level VLM planner orchestrates low-level VLA modules.

\section{High-Level VLM Prompt}
\label{sup:vlm_prompt}

\begin{tcolorbox}[
    standard jigsaw,
    title=High-level VLM Decision Making,
    opacityback=0,
    fontupper=\small,
    fontlower=\small,
    breakable
]
\setlength{\parindent}{15pt}
\noindent
\textbf{Image Input:} \textcolor{purple}{[Current Camera Observations]} \\
\textbf{Text Input:}
You are a decision-making agent tasked with generating \textit{natural language command} that is sent to a Robotics Vision-Language-Action (VLA) model for a two-armed robot towards completing the \textit{given task} \textcolor{purple}{[Task Instruction]}.
The command must be in the active, second-person voice (addressing the robot), based on the \textit{current observation} of the robot shown in the image above.
\textcolor{purple}{[Observation Representation]}.\\
Key Directives:\\
1. Output a single command that should be executed immediately. The command should facilitate completion of the given task.\\
2. The command should be doable within 10 seconds. Consider the affordance of the VLA based on the history steps as well as the current state of the robot. \\
3. Think step by step internally to arrive at the command. Do not output your thought process. \\
Current Memory: \textcolor{purple}{[Memory]}.\\\\
\textbf{VLM Policy Output: } \textcolor{purple}{[Language Command]}%\\
\end{tcolorbox}

\section{Success Detection Prompt}
\label{sup:succ_det}
\begin{tcolorbox}[
    standard jigsaw,
    title=Success Detection Prompt,
    opacityback=0,
    fontupper=\small,
    fontlower=\small,
    breakable
]
\setlength{\parindent}{15pt}
\noindent
\textbf{Image Input:} \textcolor{purple}{[A sequence of Observations]}\\\\
\textbf{Text Input:} You are a success detector for a robot. Your job is to check whether the robot has successfully completed a command
based on the current observation of the robot cameras and the scene information below.
\textcolor{purple}{[Privileged State Information]}.\\
Think about what criteria are required for success. Only output yes if all of the criteria are met.\\
Examples:\\
1) pick up requires that the object is NOT making contact with the table AND is in contact with the gripper\\
2) put in requires that the object is in contact with the container AND is NOT in contact with the gripper\\
Has the robot completed the following command "\textcolor{purple}{[Language Command]}"? Answer with only "yes" or "no" or "uncertain" (in lowercase).\\\\
\textbf{VLM Output: } \textcolor{purple}{[yes / no / uncertain]}\\
\end{tcolorbox}

\section{Observation Description Prompt}
\label{sup:obs_desc_prompt}
\begin{tcolorbox}[
    standard jigsaw,
    title=Observation Description Prompt,
    opacityback=0,
    fontupper=\small,
    fontlower=\small,
    breakable
]
\setlength{\parindent}{15pt}
\noindent
\textbf{Image Input:} \textcolor{purple}{[Current Observation]}\\\\
\textbf{Text Input:} The user wants a robot to perform the following task: \textcolor{purple}{[Instruction]}. \\
Scene information: \textcolor{purple}{[BBox info / Privileged info / None]}\\
Based on the current observation of the robot shown in the image above, and the scene information, please provide a concise description of the scene, focus only on task relevant objects and what the robot is doing (e.g. is it trying to pick something?).\\\\
\textbf{VLM Output: } \textcolor{purple}{[Scene Descriptions]}\\
\end{tcolorbox}

\section{Bounding Box Prompt}
\label{sup:bb_desc_prompt}
\begin{tcolorbox}[
    standard jigsaw,
    title=Bounding Box Prompt,
    opacityback=0,
    fontupper=\small,
    fontlower=\small,
    breakable
]
\setlength{\parindent}{15pt}
\noindent
\textbf{Image Input:} \textcolor{purple}{[Current Observation]}\\\\
\textbf{Text Input:} Detect the locations of the objects: \textcolor{purple}{[Object List]}, 'Left Gripper', 'Right Gripper'. Output a json list where each entry contains the 2D bounding box in "box\_2d" and a text label in "label". Only one entry per object.\\\\
\textbf{VLM Output:} Below are the bounding boxes for the objects in the image, which are represented as [y\_min, x\_min, y\_max, x\_max] and normalized to be between 0-1000. \textcolor{purple}{[Bounding Box JSON Output]}\\
\end{tcolorbox}

\section{Memory Summarization Prompt}
\label{sup:mem_prompt}
\begin{tcolorbox}[
    standard jigsaw,
    title=Bounding Box Prompt,
    opacityback=0,
    fontupper=\small,
    fontlower=\small,
    breakable
]
\setlength{\parindent}{15pt}
\noindent
\textbf{Input:} You are a decision-making agent tasked with generating a sequence of natural language commands for a two-armed robot to complete the given task \textcolor{purple}{[Instruction]}.\\
Step history \textcolor{blue}{[$\times n$]}:\\
\indent Step \textcolor{purple}{[i]} \\
\indent Instruction: \textcolor{purple}{[language command]}.\\
\indent Instruction successfully completed? \textcolor{purple}{[yes / no]} \\
\indent Instruction result: \textcolor{purple}{[observation representation]}

\noindent
Based on the step history above, summarize the affordance of this 
language-conditioned policy into two or three short bullet points that 
can help the agent make better decisions. Be as concise as possible.
\end{tcolorbox}

\section{Results Table}
\label{sup:results}
% Here we show the per-task results for each experiment. For each setting, we first show the aggregated results across different task types, then a detailed breakdown for each type of tasks.

% %%%%%%%%%%%%%%%%%%%%%%
\subsection{VLM}
\label{sup:vlm_graph}
\begin{table}[H]\centering
\caption{Evaluating different choices of VLM}
\label{tab:vlm}
% Set up alternating row colors starting from the second row (the data rows)
%\rowcolors{2}{white}{lightgray}
\setlength{\tabcolsep}{6pt} % Increase space between columns
\footnotesize
\begin{tabular}{l ccc}
\toprule
\textbf{VLM Configuration} & \textbf{Short-Horizon} ($\%$) & \textbf{Long-Horizon} ($\%$) & \textbf{Reasoning} ($\%$) \\
\midrule
Gemini 2.5 Flash-Lite & 70.48 $\pm$ 1.05 & 48.73 $\pm$ 1.52 & 58.51 $\pm$ 1.19\\
Gemini 2.5 Flash-Lite (thinking) & \textbf{74.44 $\pm$ 0.89} & \textbf{58.21 $\pm$ 1.52} & \textbf{75.20 $\pm$ 1.25}\\
Gemini 2.5 Flash & 72.63 $\pm$ 0.94 & 47.02 $\pm$ 1.45 & 71.79 $\pm$ 1.22\\
Gemini 2.5 Flash (thinking) & \textbf{75.81 $\pm$ 0.93} & 52.36 $\pm$ 1.54 & 72.62 $\pm$ 1.17\\
Gemini 2.5 Pro (thinking) & 70.10 $\pm$ 1.01 & 53.06 $\pm$ 1.42 & \textbf{74.39 $\pm$ 1.21}\\
\bottomrule
\end{tabular}
\end{table}

\subsection{VLA}
\label{sup:vla}

%%%%%%%%%%%%%%%%%%%%%%%%%%%%%%%%%%%%%%%
\begin{table}[H]\centering
\caption{Evaluating different low-level VLA models.}
\label{tab:vla}
%\rowcolors{2}{white}{lightgray}
\setlength{\tabcolsep}{10pt} % Increase space between columns
\footnotesize
\begin{tabular}{l ccc}
\toprule
\textbf{VLA Configuration} & \textbf{Short-Horizon} ($\%$) & \textbf{Long-Horizon} ($\%$) & \textbf{Reasoning} ($\%$) \\
\midrule
GROD-1B & 63.40 $\pm$ 1.16 & 41.30 $\pm$ 1.49 & 66.90 $\pm$ 1.25\\
GROD-1B (FT with sim) & 54.60 $\pm$ 1.09 & 7.50 $\pm$ 0.80 & 43.00 $\pm$ 1.05\\
GROD-3B & \textbf{75.81 $\pm$ 0.93} & \textbf{52.36 $\pm$ 1.54} & \textbf{72.62 $\pm$ 1.17}\\
\bottomrule
\end{tabular}
\end{table}

\subsection{Termination Condition}
\label{sup:term_graph}

%%%%%%%%%%%%%%%%%%%%%%%%%%%%%%%%%%%%%%%
\begin{table}[H]\centering
\caption{Termination Condition}
\label{tab:term}
%\rowcolors{2}{white}{lightgray}
\setlength{\tabcolsep}{10pt} % Increase space between columns
\footnotesize
\begin{tabular}{l ccc}
\toprule
\textbf{Termination Condition} & \textbf{Short-Horizon} ($\%$) & \textbf{Long-Horizon} ($\%$) & \textbf{Reasoning} ($\%$) \\
\midrule
VLM-based Horizon & 72.16 $\pm$ 0.95 & 43.50 $\pm$ 1.14 & 72.27 $\pm$ 1.14\\
Success Detector & \textbf{74.65 $\pm$ 0.92} & \textbf{57.39 $\pm$ 1.52} & \textbf{80.89 $\pm$ 1.17}\\
Fixed Horizon ($T = 400$) & \textbf{75.81 $\pm$ 0.93} & 52.36 $\pm$ 1.54 & 72.62 $\pm$ 1.17\\
\bottomrule
\end{tabular}
\end{table}

\subsection{Observation Description}
\label{sup:obs_desc_graph}

%%%%%%%%%%%%%%%%%%%%%%%%%%%%%%%%%%%%%%%%%%%

\begin{table}[H]\centering
\caption{System performance with different observation representations}
\label{tab:obs_representations}
%\rowcolors{2}{white}{lightgray}
\setlength{\tabcolsep}{6pt} % Increase space between columns
\footnotesize
\begin{tabular}{l ccc}
\toprule
\textbf{Observation Representation} & \textbf{Short-Horizon} ($\%$) & \textbf{Long-Horizon} ($\%$) &  \textbf{Reasoning} ($\%$) \\
\midrule
Image & 67.56 $\pm$ 0.98 & 38.84 $\pm$ 1.50 & 69.21 $\pm$ 1.23\\
Image + description & 67.93 $\pm$ 1.01 & 35.70 $\pm$ 1.43 & 62.77 $\pm$ 1.31\\
Image + description + bboxes & 73.94 $\pm$ 0.95 & 47.90 $\pm$ 1.67 & 68.51 $\pm$ 1.26\\
Image + description + contact info & \textbf{75.81 $\pm$ 0.93} & \textbf{52.36 $\pm$ 1.54} & \textbf{72.62 $\pm$ 1.17}\\
\bottomrule
\end{tabular}
\end{table}

\subsection{Memory Length}
\label{sup:mem_graph}

%%%%%%%%%%%%%%%%%%%%%%%%%%%%%%%%%%%%%%%
\begin{table}[H]\centering
\caption{Memory Length}
\label{tab:mem}
% Set up alternating row colors starting from the second row (the data rows)
%\rowcolors{2}{white}{lightgray} 
\setlength{\tabcolsep}{10pt} % Increase space between columns
\footnotesize
\begin{tabular}{l ccc}
\toprule
\textbf{Memory Length} & \textbf{Short-Horizon} ($\%$) & \textbf{Long-Horizon} ($\%$) & \textbf{Reasoning Tasks} ($\%$) \\
\midrule
Full Memory     & 76.53 $\pm$ 0.99 & 58.98 $\pm$ 1.61 & 72.77 $\pm$ 1.22\\
Memory Window - 5 & 76.09 $\pm$ 0.87 & 57.76 $\pm$ 1.66 & 72.20 $\pm$ 1.26\\
Memory Window - 3 & 75.81 $\pm$ 0.93 & 58.21 $\pm$ 1.52 & 72.62 $\pm$ 1.17\\
Memory Window - 1 & 76.76 $\pm$ 0.96 & 59.89 $\pm$ 1.60 & 74.27 $\pm$ 1.20\\
\bottomrule
\end{tabular}
\end{table}

\subsection{Memory Summary}
\label{sup:mem_sum_graph}

%%%%%%%%%%%%%%%%%%%%%%%%%%%%%%%%%%%%%%%
\begin{table}[H]\centering
\caption{Memory Summarization}
\label{tab:mem_sum}
% Set up alternating row colors starting from the second row (the data rows)
%\rowcolors{2}{white}{lightgray}
\setlength{\tabcolsep}{10pt} % Increase space between columns
\footnotesize
\begin{tabular}{l ccc}
\toprule
\textbf{Memory Summarization} & \textbf{Short-Horizon} ($\%$) & \textbf{Long-Horizon} ($\%$) & \textbf{Reasoning} ($\%$) \\
\midrule
No summary & 75.81 $\pm$ 0.93 & 52.36 $\pm$ 1.54 & 72.62 $\pm$ 1.17\\
Summary of last step & 74.61 $\pm$ 1.00 & 52.57 $\pm$ 1.52 & 72.82 $\pm$ 1.03\\
Summary of current episode & 71.66 $\pm$ 1.04 & 50.12 $\pm$ 1.52 & 75.72 $\pm$ 1.17\\
Summary of previous episodes & \textbf{79.45 $\pm$ 0.81} & \textbf{60.00 $\pm$ 1.50} & \textbf{80.30 $\pm$ 1.23}\\
\bottomrule
\end{tabular}
\end{table}

%%%%%%%%%%%%%%%%%%%%%%

\section{Tasks Description}
\label{sup:task}
\begin{figure*}[t]
    \centering
    \begin{subfigure}[t]{0.24\textwidth}
        \includegraphics[width=\textwidth]{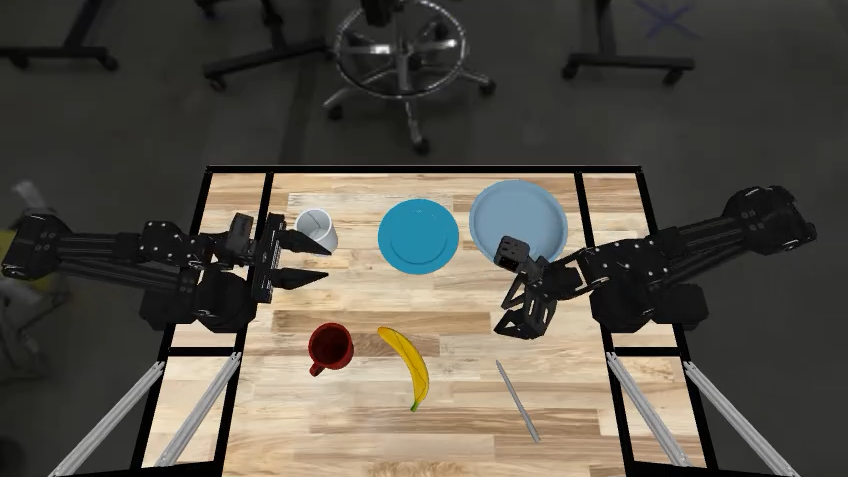}
        \caption{Dining Scene}
    \end{subfigure}
    \hfill
    \begin{subfigure}[t]{0.24\textwidth}
        \includegraphics[width=\textwidth]{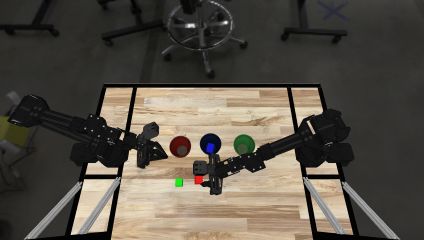}
        \caption{Cup Scene}
    \end{subfigure}
    \hfill
    \begin{subfigure}[t]{0.24\textwidth}
        \includegraphics[width=\textwidth]{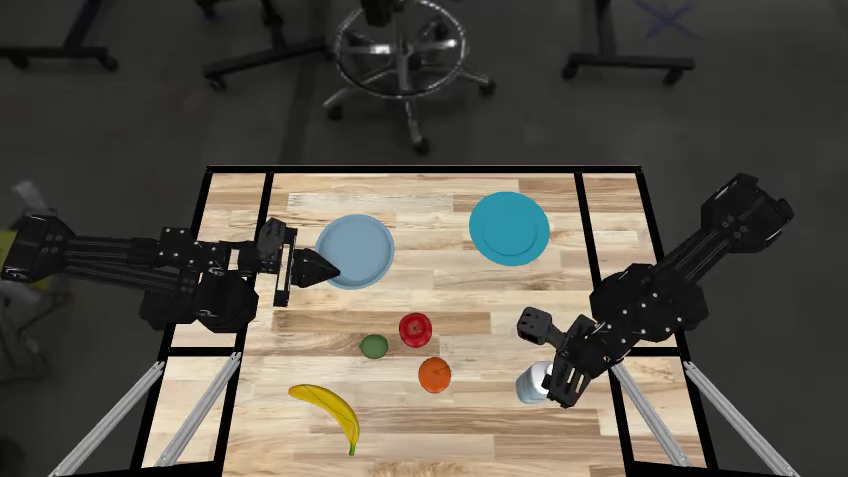}
        \caption{Fruit Scene}
    \end{subfigure}
    \hfill
    \begin{subfigure}[t]{0.24\textwidth}
        \includegraphics[width=\textwidth]{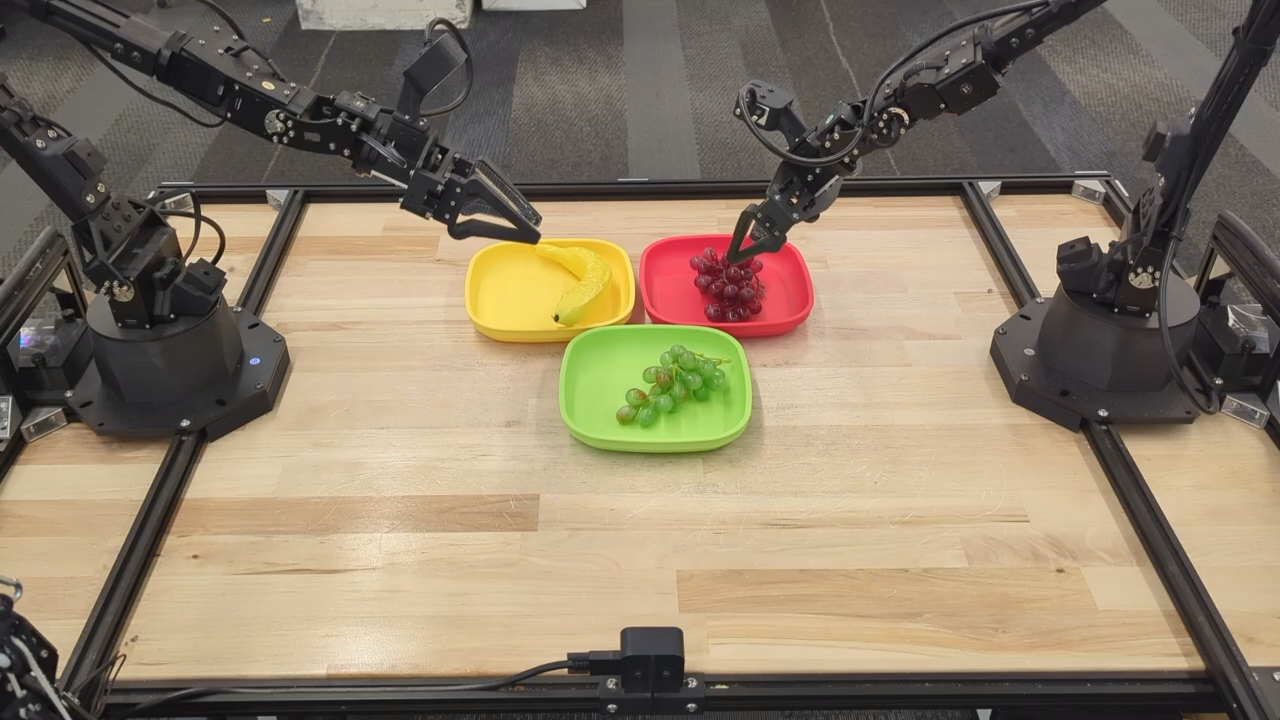}
        \caption{Real ALOHA}
    \end{subfigure}
    \caption{Example scenes from our study. Each scene is intentionally designed to support multiple different tasks, as specified in detail in Appendix~\ref{sup:task}.}
    \label{fig:four_subfigures}
\end{figure*}

In this section, we provide detailed descriptions of the tasks that we evaluated upon. 
The tasks in our main component studies are implemented based on the open-sourced MuJoCo ALOHA suite. We visualize some of these scenes as well as the real robot scene in Fig.~\ref{fig:four_subfigures}.

\subsection{Reasoning Tasks}

\textbf{Reasoning Task 1:}
\begin{itemize}
    \item \textbf{Scene}: The table contains a plate, a bowl, a white container, a red mug, a banana, and a pen.
    \item \textbf{Goal}: The robot needs to put the banana in the bowl.
    \item \textbf{Instruction:} ``Put the item that monkey can eat into the bowl''.
\end{itemize}

\vspace{0.3cm}

\noindent\textbf{Reasoning Task 2:}
\begin{itemize}
    \item \textbf{Scene}: The table contains a plate, a bowl, a white container, a red mug, a banana, and a pen.
    \item \textbf{Goal}: The robot needs to put the mug on the plate.
    \item \textbf{Instruction:} ``Put the object you pour coffee in on the plate''.
\end{itemize}

\vspace{0.3cm}

\noindent\textbf{Reasoning Task 3:}
\begin{itemize}
    \item \textbf{Scene}: The table contains three cups and three cubes, each set corresponding to the colors red, green, and blue.
    \item \textbf{Goal}: The robot needs to put each cube into the cup with the same color.
    \item \textbf{Instruction:} ``Put all the cubes into their matching colored cups.''.
\end{itemize}

\vspace{0.3cm}

\noindent\textbf{Reasoning Task 4:}
\begin{itemize}
    \item \textbf{Scene}: The table contains a plate, a bowl, a white container, a red mug, a banana, and a pen.
    \item \textbf{Goal}: The robot needs to put both the banana and the pen in the bowl.
    \item \textbf{Instruction:} ``Put all things that are long in the bowl''.
\end{itemize}

\vspace{0.3cm}

\noindent\textbf{Reasoning Task 5:}
\begin{itemize}
    \item \textbf{Scene}: The table contains a plate, a bowl, a banana, a lime, an orange, an apple, and a bottle.
    \item \textbf{Goal}: The robot needs to put the lime in the bowl.
    \item \textbf{Instruction:} ``Put the sourest fruit in the bowl.''.
\end{itemize}

\subsection{Long-horizon Tasks}

\textbf{Long-horizon Task 1:}
\begin{itemize}
    \item \textbf{Scene}: The table contains three cups and three cubes, each set corresponding to the colors red, green, and blue.
    \item \textbf{Goal}: The robot needs to put all the cubes into the green cup.
    \item \textbf{Instruction:} ``Put all the cubes into the green cup''.
\end{itemize}
\vspace{0.3cm}

\noindent\textbf{Long-horizon Task 2:}
\begin{itemize}
    \item \textbf{Scene}: The table contains three cups and three cubes, each set corresponding to the colors red, green, and blue.
    \item \textbf{Goal}: The robot needs to first stack the red cup into the green cup, and then put the red cube into the red cup.
    \item \textbf{Instruction:} ``Put the red cup into the green cup. Then put the red cube into the red cup.''.
\end{itemize}

\vspace{0.3cm}

\noindent\textbf{Long-horizon Task 3:}
\begin{itemize}
    \item \textbf{Scene}: The table contains a plate, a bowl, a white container, a red mug, a banana, and a pen.
    \item \textbf{Goal}: The robot needs to put both the banana and the pen in the bowl.

    \item \textbf{Instruction:} ``Put the banana and the pen in the bowl.''.
\end{itemize}

\vspace{0.3cm}

\noindent\textbf{Long-horizon Task 4:}
\begin{itemize}
    \item \textbf{Scene}: The table contains a plate, a bowl, a white container, a red mug, a banana, and a pen.
    \item \textbf{Goal}: The robot needs to put both the mug and the banana in the plate.
    \item \textbf{Instruction:} ``Put the banana and the mug on the plate.''.
\end{itemize}

\vspace{0.3cm}

\noindent\textbf{Long-horizon Task 5:}
\begin{itemize}
    \item \textbf{Scene}: The table contains a plate, a bowl, a white container, a red mug, a banana, and a pen.
    \item \textbf{Goal}: The robot needs to put the banana in the bowl; and also put the mug on the plate.
    \item \textbf{Instruction:} ``Put the banana in the bowl and the mug on the plate''.
\end{itemize}

\subsection{Short-horizon Tasks}

\textbf{Short-horizon Task 1:}
\begin{itemize}
    \item \textbf{Scene}: The table contains a plate, a bowl, a white container, a red mug, a banana, and a pen.
    \item \textbf{Goal}: The robot needs to put the banana in the bowl.
    \item \textbf{Instruction:} ``Put the banana in the bowl''.
\end{itemize}

\vspace{0.3cm}

\noindent\textbf{Short-horizon Task 2:}
\begin{itemize}
    \item \textbf{Scene}: The table contains a plate, a bowl, a white container, a red mug, a banana, and a pen.
    \item \textbf{Goal}: The robot needs to put the mug on the plate.
    \item \textbf{Instruction:} ``Put the mug on the plate''.
\end{itemize}

\vspace{0.3cm}

\noindent\textbf{Short-horizon Task 3:}
\begin{itemize}
    \item \textbf{Scene}: The table contains a plate, a bowl, a white container, a red mug, a banana, and a pen.
    \item \textbf{Goal}: The robot needs to put the pen in the container.
    \item \textbf{Instruction:} ``Put the pen in the white container''.
\end{itemize}

\vspace{0.3cm}

\noindent\textbf{Short-horizon Task 4:}
\begin{itemize}
    \item \textbf{Scene}: The table contains a plate, a bowl, a banana, a lime, an orange, an apple, and a bottle.
    \item \textbf{Goal}: The robot needs to put the bottle in the bowl.
    \item \textbf{Instruction:} ``Place the bottle in the left bowl''.
\end{itemize}

\vspace{0.3cm}

\noindent\textbf{Short-horizon Task 5:}
\begin{itemize}
    \item \textbf{Scene}: The table contains a plate, a bowl, a banana, a lime, an orange, an apple, and a bottle.
    \item \textbf{Goal}: The robot needs to put the lime in the bowl.
    \item \textbf{Instruction:} ``Place the lime in the left bowl''.
\end{itemize}

\subsection{Cross-Stack Validation Tasks}
\label{sup:cross_stack_tasks}

Each task requires two sequential pick-and-place operations and specifies both target objects indirectly, so the high-level VLM must translate the task into commands that the VLA can execute. In the first task, the robot is asked to first place ``the condiment you would put on a hot dog'' in the basket, and then place ``the drink you would pour over cereal'' in the basket; the corresponding objects are ketchup and milk. In the second, it must first place ``the thing you would pour over a green salad'' in the basket, and then place ``the red condiment commonly served with fries'' in the basket; the corresponding objects are salad dressing and ketchup. In the third, it must first place ``what you would drizzle on lettuce'' in the basket, and then place ``the sauce commonly brushed on grilled food'' in the basket; the corresponding objects are salad dressing and BBQ sauce. This design keeps each low-level command short while testing semantic reasoning, skill composition across two sequential manipulations, and compatibility between VLM-generated commands and the VLA's language affordance.

\end{document}